%% file: main.tex
\documentclass{article}

\usepackage[final]{neurips_2023}

\usepackage[utf8]{inputenc} 
\usepackage[T1]{fontenc}    
\usepackage{hyperref}       
\usepackage{url}            
\usepackage{booktabs}       
\usepackage{amsfonts}       
\usepackage{amsmath}
\usepackage{amssymb}
\usepackage{amsthm}
\usepackage{algorithm}
\usepackage{algorithmic}
\usepackage{nicefrac}       
\usepackage{microtype}      
\usepackage{xcolor, soul}         
\usepackage{graphicx}
\usepackage{subcaption}
\usepackage{multirow}
\usepackage{multicol}
\usepackage{mathtools}
\usepackage{wrapfig}
\usepackage[capitalize,noabbrev]{cleveref}
\usepackage{pifont}
\newcommand{\cmark}{\ding{51}}%
\newcommand{\xmark}{\ding{55}}%

\DeclareMathOperator*{\argmin}{arg\,min}
\newcommand{\methodnofed}{L2P}
\newcommand{\method}{FedL2P}
\newcommand{\metanets}{meta-nets}
\newcommand{\metanet}{meta-net}

\newcommand\Tstrut{\rule{0pt}{2.5ex}}         

\title{\method: Federated Learning to Personalize}

\usepackage{background}

\backgroundsetup{angle=0, 
scale=1,
color=black,
firstpage=true,
position=current page.north,
hshift=0pt,
vshift=-20pt,
contents={\ifnum\value{page}=1 PREPRINT: Accepted at the 37th Conference on Neural Information Processing Systems (NeurIPS 2023) \else \fi}
}

\author{%
  Royson Lee\textsuperscript{1,2}\thanks{Corresponding Author: \texttt{dsrl2@cam.ac.uk}} , Minyoung Kim\textsuperscript{2}, Da Li\textsuperscript{2} \\
  \textbf{Xinchi Qiu\textsuperscript{1}, Timothy Hospedales\textsuperscript{2,3}, Ferenc Huszár\textsuperscript{1}, Nicholas D. Lane\textsuperscript{1,4}} \\
  \\
  \textsuperscript{1} University of Cambridge, UK \hspace*{10pt}
  \textsuperscript{2} Samsung AI Center, Cambridge, UK\\
  \textsuperscript{3} University of Edinburgh, UK \hspace*{10pt}
  \textsuperscript{4} Flower Labs \\
}

\begin{document}



\maketitle

\begin{abstract}
Federated learning (FL) research has made progress in developing algorithms for distributed learning of global models, as well as algorithms for local personalization of those common models to the specifics of each client’s local data distribution. However, different FL problems may require different personalization strategies, and it may not even be possible to define an effective one-size-fits-all personalization strategy for all clients: depending on how similar each client’s optimal predictor is to that of the global model, different personalization strategies may be preferred. In this paper, we consider the federated meta-learning problem of learning personalization strategies. Specifically, we consider \metanets{} that induce the batch-norm and learning rate parameters for each client given local data statistics. By learning these \metanets{} through FL, we allow the whole FL network to collaborate in learning a customized personalization strategy for each client. Empirical results show that this framework improves on a range of standard hand-crafted personalization baselines in both label and feature shift situations.\footnote[0]{Code is available at https://github.com/royson/fedl2p}
\end{abstract}

\input{01intro.tex}

\input{02related.tex}
\input{03approach.tex}

\input{04eval.tex}
\input{05conclusion.tex}

\section*{Acknowledgements}
This work was supported by Samsung AI and the European
Research Council via the REDIAL project.

\setcitestyle{numbers}
\bibliographystyle{plainnat}
\bibliography{main}

\clearpage

\appendix
\centerline{\huge\textbf{{Appendix}}}
\input{appendix/app_ift.tex}
\input{appendix/app_usecase.tex}

\input{appendix/app_cost.tex}
\input{appendix/app_pretrain.tex}
\input{appendix/app_arch.tex}

\input{appendix/app_results.tex}

\end{document}

%% file: 01intro.tex
\section{Introduction}\label{sec:01intro}
\vspace{-0.5em}

Federated learning (FL) is an emerging approach to enable privacy-preserving collaborative learning among clients who hold their own data. A major challenge of FL is to learn from differing degrees of statistical data heterogeneity among clients. This makes it hard to reliably learn a global model and also that the global model may perform sub-optimally for each local client. These two issues are often dealt with respectively by developing robust algorithms for learning the global model~\cite{fedprox,moon,ccvr} and then offering each client the opportunity to personalize the global model to its own unique local statistics via fine-tuning. In this paper, we focus on improving the fine-tuning process by learning a personalization strategy for each client. 

A variety of approaches have been proposed for client personalization. Some algorithms directly learn the personalized models~\cite{fedfomo,fedem}, 
but the majority obtain the personalized model after global model learning by fine-tuning techniques such as basic fine-tuning~\cite{fedavg,fedbabu}, regularised fine-tuning~\cite{ditto,pfedme}, and selective parameter fine-tuning~\cite{ifca,fedrep,fedper,lgfedavg}. Recent benchmarks~\cite{matsuda2022empirical,chen2022pfl} showed that different personalized FL methods suffer from lack of comparable evaluation setups. In particular, dataset- and experiment-specific personalization strategies are often required to achieve state-of-the-art performance. Intuitively, different datasets and FL scenarios require different personalization strategies. For example, scenarios with greater or lesser heterogeneity among clients, would imply different strengths of personalization are optimal. Furthermore, exactly how that personalization should be conducted might depend on whether the heterogeneity is primarily in marginal label shift, marginal feature shift, or conditional shift. None of these facets can be well addressed by a one size fits all personalization algorithm. 
Furthermore, we identify a previously understudied issue: even for a single federated learning scenario, heterogeneous clients may require different personalization strategies. For example, the optimal personalization strategy will have client-wise dependence on whether that client is more or less similar to the global model in either marginal or conditional data distributions. Existing works that learn personalized strategies through the use of personalized weights are not scalable to larger setups and models~\cite{pfedhn,apfl}. On the other hand, the few studies that attempt to optimize hyperparameters for fine-tuning do not sufficiently address this issue as they either learn a single set of personalization hyperparameters~\cite{flora,holly2022evaluation} and/or learn a hyperparameter distribution without taking account of the client's data distribution~\cite{fedex}. 

In this paper, we address the issues above by considering the challenge of federated meta-learning of personalization strategies. Rather than manually defining a personalization strategy as is mainstream in FL~\cite{jiang2019improving,chen2022pfl}, we use hyper-gradient meta-learning strategies to efficiently estimate personalized 
hyperparameters. However, 
apart from standard centralised meta-learning and hyperparameter optimization (HPO) studies which only need to learn a single set of hyperparameters, we learn \metanets{} which inductively map from local client statistics to client-specific personalization hyperparameters. More specifically, our approach \method{} introduces \metanets{} to estimate the extent in which to utilize the client-wise BN statistics as opposed to the global model's BN statistics, as well as to infer layer-wise learning rates given each client’s metadata. 

Our \method{} thus enables per-dataset/scenario, as well as per-client, personalization strategy learning. By conducting federated meta-learning of the personalization hyperparameter networks, we simultaneously allow each client to benefit from its own personalization strategy, (e.g., learning 
rapidly, depending on similarity to the global model), and also  enable all the clients to collaborate by learning the overall hyperparameter networks that map local meta-data to local personalization strategy. Our \method{} generalizes many existing frameworks as special cases, such as FedBN~\cite{fedbn}, which makes a manual choice to normalize features using the client BN's statistics, and various selective fine-tuning approaches~\cite{lgfedavg,fedbabu,fedrep,fedper}, which make manual choices on which layers to personalize.

%% file: 02related.tex
\section{Related Work}\label{sec:02related}
\vspace{-1.0em}
Existing FL works aim to tackle the statistical heterogeneity of learning personalized models by either first learning a global model~\cite{fedavg,ccvr,moon,fedbabu,scaffold} and then fine-tuning it on local data or directly learning the local models, which can often be further personalized using fine-tuning.
Many personalized FL approaches include the use of transfer learning between global~\cite{fml} and local models~\cite{fedme}, model regularization~\cite{ditto}, Moreau envelopes~\cite{pfedme}, and meta-learning~\cite{perfedavg,fedmeta}. Besides algorithmic changes, many works also proposed model decoupling, in which layers are either shared or personalized~\cite{fedper,fedrep,feddar,fedbabu,lgfedavg, ifca, perfedmask} or client clustering, which assumes a local model for each cluster~\cite{hypcluster,fedme,cfl,flhc}. 
These methods often rely on or adopt a fixed personalization policy for local fine-tuning in order to adapt a global model or further improve personalized performance. 
Although there exists numerous FL approaches that propose adaptable personalization policies~\cite{pfedhn,pFedLA,apfl}, these works are memory intensive and do not scale to larger setups and models. 
On the other hand, our approach has a low memory footprint (Appendix~\ref{app:cost}) and is directly applicable and complementary to existing FL approaches as it aims to solely improve the fine-tuning process.

Another line of work involves 
HPO for FL (or FL for HPO). 
These methods either learn one set of hyperparameters for all clients~\cite{holly2022evaluation,fedex,flora} or random sample from learnt hyperparameter categorical distributions which does not take into account of the client's meta-data~\cite{fedex}.
Moreover, some of these methods~\cite{holly2022evaluation,flora} search for a set of hyperparameters based on the local validation loss given the initial set of weights prior to the federated learning of the model~\cite{holly2022evaluation,flora}, which might be an inaccurate proxy to the final performance.
Unlike previous works which directly learn hyperparameters, we deploy FL to learn \metanets{} that take in, as inputs, the client meta-data to generate personalized hyperparameters for a given pretrained model. A detailed positioning of our work in comparison with existing literature can be found in Appendix~\ref{app:usecase}.

%% file: 03approach.tex
\section{Proposed Method}\label{sec:approach}
\vspace{-0.5em}
\subsection{Background \& Preliminaries}
\vspace{-0.5em}
\textbf{Centralized FL.}
A typical centralized FL setup using FedAvg~\cite{fedavg} involves training a global model $\theta_g$ from $C$ clients whose data are kept private. At round $t$, $\theta_g^{t-1}$ is broadcast to a subset of clients selected, $\tilde{C}$, using a fraction ratio $r$. Each selected client, $i \in \tilde{C}$, would then update the model using a set of hyperparameters and its own data samples, which are drawn from its local distribution $P_i$ defined over $\mathcal{X} \times \mathcal{Y}$ for a compact space $\mathcal{X}$ and a label space $ \mathcal{Y}$ for $e$ epochs. After which, the learned local models, $\theta_i^{t-1}$, are sent back to the server for aggregation $\theta_g^{t}=\sum^{\tilde{C}}_{i} \frac{N_i}{\sum_{i'} N_{i'}} \theta_i^{t-1}$ where $N_i$ is the total number of local data samples in client $i$ and the resulting $\theta_g^{t}$ is used for the next round. The aim of FL is either to minimize the global objective $\mathbb{E}_{(x,y) \sim P}\mathcal{L}(\theta_g;x,y)$ for the global data distribution $P$ or local objective $\mathbb{E}_{(x,y) \sim P_i}\mathcal{L}_i(\theta_i;x,y)$ for all $i \in C$ where $\mathcal{L}_i(\theta;x,y)$ is the loss given the model parameters $\theta$ at data point $(x,y)$. As fine-tuning is the dominant approach to either personalize from a high-performing $\theta_g$ or to optimize $\theta_i$ further for each client, achieving competitive or state-of-the-art results in recent benchmarks~\cite{matsuda2022empirical,chen2022pfl}, we focus on the collaborative learning of \metanets{} which generates a personalized set of hyperparameters used during fine-tuning to further improve personalized performance without compromising global performance.

\textbf{Non-IID Problem Setup.}
Unlike many previous works, our method aims to handle both common label and feature distribution shift across clients. Specifically, given features $x$ and labels $y$, we can rewrite the joint probability $P_i(x,y)$ as $P_i(x|y)P_i(y)$ and $P_i(y|x)P_i(x)$ following~\cite{kairouz2021advances}. We focus on three 
data heterogeneity settings found in many realistic settings: both label \& distribution skew in which the marginal distributions $P_i(y)$ \& $P_i(x)$ may vary across clients, respectively, and concept drift in which the conditional distribution $P_i(x|y)$ may vary across clients. 

\subsection{\method: 
FL of Personalization Strategies}
\vspace{-0.5em}

\begin{figure}[t]
\vspace{-3.5em}
    \small
    \centering
    \includegraphics[width=1.0\columnwidth]{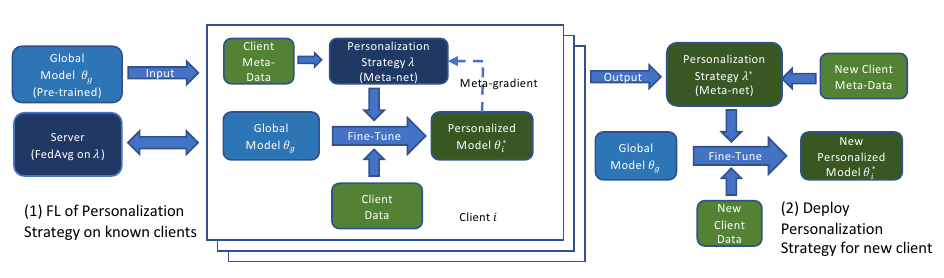}
    \captionsetup{font=small,labelfont=bf}
    \vspace{-1.5em}
    \caption{\method{}'s workflow. For each client, the meta-net (parameterized by $\lambda$) takes the client meta-data (e.g., local data profile) and outputs hyperparameters. We update $\lambda$ by optimizing the meta objective which is the validation loss of the model finetuned with the hyperparameters returned by the meta-net. The updated personalization strategies $\lambda^*$ from clients are collected/aggregated (via FedAvg) in the server for the next round. 
    }
    \label{fig:workflow}
\vspace{-1.5em}
\end{figure}


We now present our proposed method, \method{}, to tackle collaborative learning of personalization strategies under data heterogeneity. Our main motivation is that the choice of the hyperparameters for the personalized fine-tuning, such as the learning rates and feature mean and standard deviations (SD) statistics of the batch normalization~\cite{ioffe2015batch} (BN) layers, is very crucial. Although existing FL-HPO approaches aim to learn these hyperparameters {\em directly}\footnote{To our knowledge, no previous FL-HPO works learn BN statistics, 
crucial for dealing with feature shifts.}~\cite{holly2022evaluation,flora,fedex}, we aim to do it in a meta learning or hypernetwork-like fashion: learning a neural network (dubbed {\em \metanet{}}) that takes certain profiles of client data (e.g., summary statistics of personalized data) as input and outputs near-optimal hyperparameters. The \metanet{} is learned collaboratively in an FL manner without sharing local  data. 
The returned hyperparameters from the meta-net are then deployed in client's personalized fine-tuning. 
The main advantage of this \metanet{} approach over the direct HPO is that a brand new client needs not do time-consuming HPO, but just gets their optimal hyperpameters by a single feed-forward pass through the meta-net. 
Our idea is visualized in Fig.~\ref{fig:workflow}. 
For each FL round, the latest \metanet{} is distributed to the participating clients; each client then performs meta learning to update the \metanet{}; the local updated \metanets{} are sent to the server for aggregation. Details of our algorithm are described below.

\textbf{Hyperparameter Selection.}\quad 
There is a wide range of local training hyperparameters that can be optimized, some of which were explored in previous FL works~\cite{holly2022evaluation,flora,fedex}. 
As local training is often a costly process, we narrowed it down to two main sets of hyperparameters based on previous works that showed promising results in dealing with non-IID data: {\em BN hyperparameters} and {\em selective update hyperparameters}.

\textbf{Batch Normalization Hyperparameters.}\quad 
The first set of hyperparameters involves BN and explicitly deals with feature shift. Notably, FedBN~\cite{fedbn} proposed keeping the BN layers local and the other layers global to better handle feature shift across clients; BN has found success at mitigating domain shifts in various domain adaptation tasks~\cite{li2016revisiting,chang2019domain} and can be formulated as follows:
\begin{equation}\label{eq:bn}
g(x) = \frac{x - \hat{\mu}}{\sqrt{\hat{\sigma}^2 + \epsilon}} * \gamma + \delta
\end{equation}
where $g$ is a BN layer, $x$ is its input features, ($\hat{\mu}$, $\hat{\sigma}^2$) are the estimated running mean, variance, and $\epsilon$ is used for numerical stability. During training, the batch statistics $E(x)$ and $Var(x)$ are used instead, keeping running estimates of them, $\hat{\mu}$ and $\hat{\sigma}^2$. During inference, these estimates are used\footnote{When the running mean $\&$ variance are not tracked, the batch statistics is used in both training and inference.}. Both $\gamma$ and $\delta$ are learnable parameters used to scale and shift the normalized features.

Although deploying local BN layers is useful to counteract the drawbacks of feature shift, using global BN layers is often beneficial for local fine-tuning in cases where the feature shift is minimal as it helps speed-up convergence and reduces overfitting. Hence, we propose learning a hyperparameter $\beta$ for each BN layer as follows:
\begin{align}
\vspace{-1.0em}
\begin{split}\label{eq:alpha}
\hat{\mu} = (1 - \beta) * \hat{\mu}_{pt} + \beta * \hat{\mu}_{i},  \ \ \ \ \ \ \ \
\hat{\sigma}^2 = (1 - \beta) * \hat{\sigma}^2_{pt} + \beta * \hat{\sigma}^2_{i}
\end{split}
\end{align}
where $\hat{\mu}_{pt}$ and  $\hat{\sigma}^2_{pt}$ are the estimated running mean and variance by the given pretrained model; $\hat{\mu}_{i}$ and  $\hat{\sigma}^2_{i}$ are the running mean and variance estimated by the client. When $\beta\!\rightarrow\!0$, the client solely uses pretrained model's BN statistics and when $\beta\!\rightarrow\!1$, it uses its local statistics to normalize features. Thus $\beta$ indicates the degree in which the client should utilize its own BN statistics against pretrained model's, to handle the feature shift in its local data distribution.

\input{algorithms/fedft.tex}

\textbf{Selective Update Hyperparameters.}\quad A variety of recent personalized FL works achieved promising results by manually selecting and fine-tuning a sub-module of $\theta_g$ during personalization~\cite{fedbabu,fedper,lgfedavg}, (\textit{e.g.}, the feature extractor or the classifier layers), while leaving the remaining modules in $\theta_g$ frozen. It is beneficial because it allows the designer to manage over- vs under-fitting in personalization. \textit{e.g.}, if the per-client dataset is small, then fine-tuning many parameters can easily lead to overfitting, and thus better freezing some layers during personalization. Alternatively, if the clients differ significantly from the global model and/or if the per-client dataset is larger, then more layers could be beneficially personalised without underfitting. Clearly the optimal configuration for allowing model updates depends on the specific scenario and the specific client. Furthermore, it may be beneficial to consider a more flexible range of hyper-parameters that control a continuous degree of fine-tuning strength, rather than a binary frozen/updated decision per module. 

To automate the search for good personalization strategies covering a range of wider and more challenging non-IID setups, we consider  layer-wise learning rates, $\eta$ for all learnable weights and biases. This parameterization of personalization encompasses all the previous manual frozen/updated split approaches as special cases. Furthermore, while these approaches have primarily considered heterogeneity in the marginal label distribution, we also aim to cover feature distribution shift between clients. Thus we also include the learning rates for the BN parameters, $\gamma$ and $\delta$, allowing us to further tackle feature shift by adjusting the means and SD of the normalized features. 

\textbf{Hyperparameter Inference for Personalization.}\quad
We aim to estimate a set of local hyperparameters that can best personalize the pretrained model for each client given a group of clients whose data might not be used for pretraining. To accomplish this, we learn to estimate hyperparameters based on the degree of data heterogeneity of the client's local data with respect to the data that the model is pretrained on. 
There are many ways to quantify data heterogeneity, such as utilizing the earth mover's distance between the client's data distribution and the population distribution for label distribution skew~\cite{emd} or taking the difference in local covariance among clients for feature shift~\cite{fedbn}. In our case, we aim to distinguish both label and feature data heterogeneity across clients. To this end, we utilize the client's local input features to each layer with respect to the given pretrained model.
Given a pretrained model with $M$ layers and $B$ BN layers, we learn $\boldsymbol{\eta}=\eta_1 ,\hdots, \eta_{2M}$\footnote{We assume all layers have weights and biases here.} and $\boldsymbol{\beta}=\beta_1, \hdots, \beta_B$, using two functions, each of which is parameterized as a multilayer perceptron (MLP), named \metanet{}, with one hidden layer due to its ability to theoretically approximate almost any continuous function~\cite{csaji2001approximation,meta_weight_net}. We named the \metanet{} that estimates $\boldsymbol{\beta}$ and the \metanet{} that estimates $\boldsymbol{\eta}$ as BNNet and LRNet respectively. Details about the architecture can be found in Appendix.~\ref{app:arch}. 

To estimate $\boldsymbol{\beta}$, we first perform a forward pass of the local dataset on the given pretrained model, computing the mean and SD of each channel of each input feature for each BN layer. We then measure the distance between the local feature distributions and the pretrained model's running estimated feature distributions of the $b$-th BN layer as follows:
\begin{align}
\xi_{i,b} = \frac{1}{J} \sum\nolimits^J_{j=1} \frac{1}{2} \Big( D_{KL}(P_j||Q_j) + D_{KL}(Q_j||P_j) \Big)
\label{eq:kld},
\end{align}
where $P_j = \mathcal{N}(\mu_{i,b,j},\sigma^2_{i,b,j})$,
$Q_j = \mathcal{N}(\hat{\mu}_{pt,b,j},\hat{\sigma}^2_{pt,b,j})$, $D_{KL}$ is the 
KL divergence and $J$ is the number of channels of the input feature. $\xi$ is then used as an input to BNNet, which learns to estimate $\boldsymbol{\beta}$ as shown in Eq.~\ref{eq:est}. 

Similarly, we compute the mean and SD of each input feature per layer by performing a forward pass of the local dataset on the pretrained model and use it as an input to LRNet. Following best practices from previous non-FL hyperparameter optimization works~\cite{oreshkin2018tadam,baik2020meta}, we use a learnable post-multiplier $\tilde{\boldsymbol{\eta}}=\tilde{\eta}_1, \hdots, \tilde{\eta}_{2M}$ to avoid limiting the range of the resulting learning rates (Eq~\ref{eq:est}). 
\begin{align}\label{eq:est}
\begin{split}
\boldsymbol{\beta} = \text{BNNet}&(\boldsymbol{w}_{bn};\xi_1,\xi_2,\hdots,\xi_{B-1},\xi_{B}) \\
\boldsymbol{\eta} = \text{LRNet}&(\boldsymbol{w}_{lr};E(x_0),SD(x_0),E(x_1),SD(x_1)
\hdots,E(x_{M-1}),SD(x_{M-1})) \odot \tilde{\boldsymbol{\eta}}
\end{split}
\end{align}
where $\odot$ is the Hadamard product, $x_{m-1}$ is the input feature to the $m$-th layer, and $\boldsymbol{w}_{bn}$ and $\boldsymbol{w}_{lr}$ are the parameters of BNNet and LRNet respectively. $\boldsymbol{\beta}$ is used to compute the running mean and variance in the forward pass for each BN layer as shown in Eq.~\ref{eq:alpha} and $\boldsymbol{\eta}$ is used as the learning rate for each weight and bias in the backward pass. We do not restrict $\tilde{\boldsymbol{\eta}}$ to be positive as the optimal learning rate might be negative~\cite{bernacchia2021metalearning}.

\textbf{Federated Hyperparameter Learning.}
We deploy FedAvg~\cite{fedavg} to federatedly learn a set of client-specific personalization strategies. Specifically, we learn the common \metanet{} $\lambda=\{\boldsymbol{w}_{bn},\boldsymbol{w}_{lr},\tilde{\boldsymbol{\eta}}\}$ that generates client-wise personalization hyperparameters $\{{\boldsymbol{\beta}_i,\boldsymbol{\eta}_i\}}$, such that a group of clients can better adapt a pre-trained model $\theta_g$ by fine-tuning to their local data distribution. So we solve:
\begin{align}
\vspace{-2.5em}
\min_{\lambda} \mathcal{F}(\lambda,\theta_g) &= \sum^C_{i=1} \frac{N_i}{\sum_{i'} N_{i'}} \mathcal{L}_{i,V}(\theta_i^*(\lambda),\lambda) \;\; \nonumber \\
\text{s.t.}  \;\; 
\theta_i^*(\lambda) &= \argmin_{\theta_i} \mathcal{L}_{i,T}(\theta_i,\lambda) \label{eq:obj}
\vspace{-2.5em}
\end{align}
where $\theta_i^*$ is the set of optimal personalized model parameters after fine-tuning $\theta_g$ for $e$ epochs on the local dataset, $\mathcal{L}_{i,V}(\theta,\lambda) = \mathbb{E}_{(x,y) \sim V_i} \mathcal{L}_{i}(\theta,\lambda; x, y)$ and $V_i$ is the validation set (samples from $P_i$) for the client $i$ - similarity for $L_{i,T}$.  

For each client $i$, the validation loss gradient with respect to $\lambda$, known as the hypergradient, can be computed as follows:
\begin{align}\label{eq:grad_lambda}
\begin{split}
d_\lambda \mathcal{L}_{V}(\theta^*(\lambda), \lambda) = \partial_\lambda \mathcal{L}_{V}(\theta^*(\lambda),\lambda) + \partial_{\theta^*(\lambda)} \mathcal{L}_{V}(\theta^*(\lambda),\lambda) \:
\partial_\lambda \theta^*(\lambda)  
\end{split}
\end{align}
To compute $\partial_\lambda \theta^*$ in Eq.~\ref{eq:grad_lambda}, we use the implicit function theorem (IFT):
\vspace{-0.7mm}
\begin{equation}\label{eq:ift}
\partial_\lambda \theta^* |_{\lambda^{'}} = -(\partial_\theta^2\mathcal{L}_{T}(\theta,\lambda))^{-1} \: \partial_{\lambda\theta} \mathcal{L}_{T}(\theta,\lambda) |_{\lambda^{'},\theta^*(\lambda^{'})}
\end{equation}
The full derivation is shown in Appendix \ref{app:ift}.
We use Neumann approximation and efficient vector-Jacobian product as proposed by~\citet{lorraine2020optimizing} to approximate the Hessian inverse in Eq.~\ref{eq:ift} and compute the hypergradient, which is further summarized in Algorithm~\ref{algo:hypergrad}. In practice, $\theta^*$ is approximated by fine-tuning $\theta_g$ on $\mathcal{L}_T$ using the client's dataset.
Note that unlike in many previous works~\cite{lorraine2020optimizing,navon2021auxiliary} where $\partial_\lambda \mathcal{L}_{V}$ is often $0$ as the hyperparameters often do not directly affect the validation loss, in our case $\partial_{\boldsymbol{w}_{bn}} \mathcal{L}_{V} \neq 0$.


Algorithm~\ref{algo:fedft} summarizes \method{}. Given a pretrained model and a new group of clients to personalize, we first initialize $\lambda$ (line 1). For every FL round, we sample $Cr$ clients and send both the parameters of the pretrained model and $\lambda$ to each client (lines 3-5). Each client then performs a forward pass of their local dataset to compute the mean (E) and standard deviation (SD) of the input features to each layer and the statistical distance between the local feature distributions and the pretrained model's running estimated feature distributions for each BN layer (lines 6-7). $\lambda$ is then trained for $K$ iterations; each iteration optimizes the pretrained model on $\mathcal{L}_T$ for $e$ epochs, applying $\boldsymbol{\beta}$ and $\boldsymbol{\eta}$ computed using Eq.~\ref{eq:est} (lines 8-9) at every forward and backward pass respectively. Each client then computes the hypergradient of $\lambda$ as per Algorithm.~\ref{algo:hypergrad} and update $\lambda$ at the end of every iteration (line 10). Finally, after $K$ iterations, each client sends back the updated $\lambda$ and its number of data samples, which is used for aggregation using FedAvg~\cite{fedavg} (lines 12-14). The resulting $\lambda$ is then used for personalization: each client finetunes the model using its training set and evaluates it using its test set.

\vspace{-2mm}
\subsection{Adapting the Losses for IFT}

In the IFT, we solve the following problem:
\begin{align}
\min_{\lambda} \ \mathcal{L}_V(\theta^*(\lambda), \lambda) 
\ \ \textrm{s.t.} \ \ \theta^*(\lambda) = \arg\min_\theta \mathcal{L}_T(\theta,\lambda). \label{eq:ift_inner}
\end{align}
For the current $\lambda$, we first find $\theta^*(\lambda)$ in (\ref{eq:ift_inner}) by performing several SGD steps with the training loss $\mathcal{L}_T$. Once $\theta^*(\lambda)$ is obtained, we can compute the hypergradient $d_\lambda \mathcal{L}_V(\theta^*(\lambda),\lambda)$ by the IFT, which is used for updating $\lambda$. As described in (\ref{eq:grad_lambda}) and (\ref{eq:ift}), this hypergradient requires $\partial_\lambda\mathcal{L}_T(\theta,\lambda)$, implying that the training loss has to be explicitly dependent on the hyperparameter $\lambda$. As alluded in~\citet{lorraine2020optimizing}, it is usually not straightforward to optimize the learning rate hyperparameter via the IFT, mainly due to the difficulty of expressing the dependency of the training loss on learning rates. To address this issue, we define the training loss as follows:
\begin{align}
&\mathcal{L}_T(\theta,\lambda) = \mathbb{E}_{(x,y)\sim P_T} CE(f_{\theta',{\boldsymbol\beta}(\lambda)}(x),y) \ \ \textrm{where} \label{eq:ift_loss_train-1} \\
&\ \ \ \ \theta' = \theta - {\boldsymbol\eta}(\lambda) \nabla_\theta \mathbb{E}_{(x,y)\sim P_T} CE(f_{\theta,{\boldsymbol\beta}(\lambda)}(x),y). \label{eq:ift_loss_train-2}
\end{align}
Here $f_{\theta,{\boldsymbol\beta}}(x)$ indicates the forward pass with network weights $\theta$ and the batch norm statistics ${\boldsymbol\beta}$, and $CE()$ is the cross-entropy loss. Note that in (\ref{eq:ift_loss_train-2}), we can take several (not just one) gradient update steps to obtain $\theta'$.
Now, we can see that $\mathcal{L}_T(\theta, \lambda)$ defined as above has explicit dependency on the learning rates ${\boldsymbol\eta}(\lambda)$. Interestingly, the stationary point $\theta^*(\lambda)$ of $\mathcal{L}_T(\theta,\lambda)$ coincides with $\theta'$, that is, $\theta^*(\lambda)=\theta'$, which allows for a single instance of inner-loop iterations as Line 9 in Alg.~\ref{algo:fedft}. 
Finally, the validation loss is defined as:
\begin{align}
\mathcal{L}_V(\theta,\lambda)\!=\!\mathbb{E}_{(x,y)\sim P_V} CE(f_{\theta,{\boldsymbol\beta}(\lambda)}(x),y), 
\nonumber
\end{align}
showing clear dependency on BNNet parameters through ${\boldsymbol\beta}(\lambda)$ as discussed in the previous section.

%% file: algorithms/fedft.tex
\begin{wrapfigure}{R}{0.52\textwidth}
\vspace{-2.2em}
\begin{minipage}[t]{0.52\textwidth}
\begin{algorithm}[H]
    \small
    \caption{\small{\method: FL of \metanets~for Personalization Hyperparameters}} \label{algo:fedft}
    \algsetup{indent=0.5em}
    \textbf{Input:} \footnotesize{Pretrained global model $\theta_g$ with $M$ layers and $B$ BN layers, each has $J$ input channels. Fraction ratio $r$. Total no. of clients $C$. No. of update iterations $K$. Training loss $\mathcal{L}_T$ and validation loss $\mathcal{L}_V$. $\zeta$ is the learning rate for $\lambda$.}
    \begin{algorithmic}[1]
        \STATE initialize $\lambda=\{\boldsymbol{w}_{bn},\boldsymbol{w}_{lr},\tilde{\boldsymbol{\eta}}\}$
        \FOR{round $t = 1,\hdots,T$}
            \STATE $\tilde{C}$ $\leftarrow$ Random sample $Cr$ clients
            \FOR{client $i \in \tilde{C}$}
                \STATE send $\theta_g, \lambda$ to client
                \STATE Forward pass of local dataset to compute \\ 1) $E(x_m),SD(x_m)$ for $1 \leq m \leq M-1$ \\ 2) $\mu_{b,j},\sigma^2_{b,j}$ for $b=1,\hdots,B$ and $j=1,\hdots,J$.
                \STATE Compute $\xi_b$ for $b=1,\hdots,B$ using Eq.~\ref{eq:kld}
                \FOR{iteration $k=1,\hdots,K$}
                    \STATE $\theta_i \leftarrow $ Finetune $\theta_g$ using $\mathcal{L}_T$ for $e$ epochs
                    \STATE $\lambda \leftarrow \lambda - \zeta \:$ \textit{Hypergradient}($\mathcal{L}_V$, $\mathcal{L}_T$, $\lambda$, $\theta_i$)
                \ENDFOR
                \STATE send $\lambda$ and num of data samples $N$ to server
            \ENDFOR
            \STATE $\lambda \leftarrow \sum^{\tilde{C}}_{i} \frac{N_i}{\sum N_i} \lambda_i$ 
        \ENDFOR
    \end{algorithmic}
    \textbf{Output:} $\lambda$  
\end{algorithm}
\vspace{-2em}
\end{minipage}
\begin{minipage}[t]{0.52\textwidth}
\begin{algorithm}[H]
    \small
    \caption{\small{Hypergradient}} \label{algo:hypergrad}
    \algsetup{indent=0.5em}
    \vspace{-0.4em}
    \begin{flushleft}
    \textbf{Input:} \footnotesize{Validation Loss $\mathcal{L}_V$ and training Loss $\mathcal{L}_T$. Learning rate $\psi$ and no. of iterations $Q$.
    Fixed point $(\lambda^{'}, \theta^*(\lambda^{'}))$}.
    \end{flushleft}
    \begin{algorithmic}[1]
    \STATE $p = v = \partial_{\theta} \mathcal{L}_V |_{(\lambda^{'}, \theta^*(\lambda^{'}))}$
    \FOR{iteration $1,\hdots,Q$}
        \STATE $v \leftarrow v - \psi \: v \: \partial^2_{\theta^*}\mathcal{L}_T$
        \STATE $p \leftarrow p + v$
    \ENDFOR
    \end{algorithmic}
    
    \textbf{Output: $\partial_\lambda \mathcal{L}_{V} |_{(\lambda^{'}, \theta^*(\lambda^{'}))} - p \: \partial_{\lambda\theta} \mathcal{L}_T |_{(\lambda^{'}, \theta^*(\lambda^{'}))}$} 
\end{algorithm}
\vspace{-3em}
\end{minipage}
\end{wrapfigure}

%% file: 04eval.tex
\section{Evaluation}\label{sec:04eval}
\vspace{-0.5em}

\input{tables/cifar_extended.tex}

\subsection{Experimental Setup}
\vspace{-0.5em}
Experiments are conducted on image classification tasks of different complexity.
We use ResNet-18~\cite{resnet} for all experiments and SGD for all optimizers. All details of the pretrained models can be found in Appendix.~\ref{app:pretrain}.
Additionally, the batch size is set to $32$ and the number of local epochs, $e$, is set to $15$ unless stated otherwise. 
The learning rate ($\zeta$) for $\lambda=\{\boldsymbol{w}_{bn},\boldsymbol{w}_{lr},\tilde{\boldsymbol{\eta}}\}$ is set to \{$10^{-3}$,$10^{-3}$,$10^{-4}\}$, respectively. 
The hypergradient is clipped by value $[-1,1]$, $Q=3$, and $\psi=0.1$ in Alg.~\ref{algo:hypergrad}. 
The maximum number of communication rounds is set to $500$, and over the rounds we save the $\lambda$ value that leads to the lowest validation loss, averaged over the participating clients, as the final learned $\lambda$. The fraction ratio $r\!=\!0.1$ unless stated otherwise, sampling $10\%$ of the total number of clients per FL round.
We focus on non-IID labels and feature shifts while assuming that each client has an equal number of samples. 
Finally, to generate heterogeneity in label distributions, we follow the latent Dirichlet allocation (LDA) partition method~\citep{hsu2019measuring, yurochkin2019bayesian,zerofl}: $y \sim Dir(\alpha)$ for each client. 
Hence, the degree of heterogeneity in label distributions is controlled by $\alpha$; as $\alpha$ decreases, the label non-IIDness increases, and vice versa. 

\vspace{-1em}
\subsubsection{Datasets}


\textbf{CIFAR10~\cite{krizhevsky2009learning}.} A widely-used image classification dataset, also popular as an FL benchmark. The number of clients $C$ is set to $1000$ and $20\%$ of the training data is used for validation.

\textbf{CIFAR-10-C~\cite{cifar10c}.} The test split of the CIFAR10 dataset is  corrupted with common corruptions. We used $10$ corruption types\footnote{\textit{brightness}, \textit{frost}, \textit{jpeg\_compression}, \textit{contrast}, \textit{snow}, \textit{motion\_blur}, \textit{pixelate}, \textit{speckle\_noise}, \textit{fog}, \textit{saturate}} with severity level $3$ and split each corruption dataset into $80\%/20\%$ training/test sets. A subset ($20\%$) of the train set is further held out to form a validation set. Each corruption type is partitioned by $Dir(\alpha)$ among $25$ clients, hence $C=250$. 

\textbf{Office-Caltech-10~\cite{office} \& DomainNet~\cite{domainnet}.} These datasets 
are specifically designed to 
contain several domains that exhibit different feature shifts. We set the no. of samples of each domain to be the smallest of the domains, random sampling the larger domains. For Caltech-10, we set $r=1.0$ and $C=4$, one for each domain and thus partitioned labels are IID. For DomainNet, we set $C=150$, of which each of the $6$ domain datasets are partitioned by $Dir(\alpha)$ among $25$ clients, resulting in a challenging setup with both feature \& label shifts. Following FedBN~\cite{fedbn}, we take the top 10 most commonly used classes in DomainNet. 

\textbf{Speech Commands V2~\cite{warden2018speech}.} We use the 12-class version that is naturally partitioned by speaker, with one client per speaker. It is naturally imbalanced in skew and frequency, with 2112, 256, 250 clients/speakers for train, validation, and test. We sampled 250 of 2112 training clients with the most data for our \textit{seen} pool of clients and sampled 50 out of 256+250 validation and test clients for our \textit{unseen} client pool. Each client’s data is then split 80\%/20\% for train and test sets, with a further 20\% of the resulting train set held out to form a validation set.

\vspace{-1em}
\subsubsection{Baselines}\label{subsec:baseline}
\vspace{-0.5em}
We run all experiments three times and report the mean and SD of the test accuracy. We consider three different setups of fine-tuning in our experiments as below. As basic fine-tuning (FT) is used ubiquitously in FL, our experiments directly compare FT with \method{} in different non-IID scenarios. We are agnostic to the FL method used to obtain the global model or to train the \metanets{}. 

\textbf{FT (BN C).} Client BN Statistics. Equivalent to setting $\beta=1$ in  Eq.~\ref{eq:alpha}, thus using client statistics to normalize features during fine-tuning. This is similar to FedBN~\cite{fedbn}, and is adopted by FL works such as Ditto~\cite{ditto} and PerFedAvg~\cite{perfedavg}. 
\vspace{-0.5em}

\textbf{FT (BN G).} Equivalent to setting $\beta=0$ in Eq.~\ref{eq:alpha}. BN layers use the pretrained global model's BN statistics to normalize its features during fine-tuning. 
\vspace{-0.5em}

\textbf{FT (BN I).} BN layers use the incoming feature batch statistics to normalize its features during fine-tuning. This setting is adopted by FL works such as FedBABU~\cite{fedbabu}.
\vspace{-0.5em}

\textbf{\methodnofed.} Our proposed method without FL; $\lambda$ is learnt locally given the same compute budget before being used for personalization. Hence, \methodnofed{} does client-wise HPO independently.

\subsection{Experiments on Marginal Label Shift}\label{subsec:cifar}
\vspace{-0.5em}
\input{tables/cifar.tex}
We evaluate our approach based on the conventional setup of personalized FL approaches~\cite{chen2022pfl,matsuda2022empirical}, where a global model, $\theta_g$, is first learned federatedly using existing algorithms 
and then personalized to the same set of clients via fine-tuning. To this end, given the CIFAR-10 dataset partitioned among a group of clients, we pretrained $\theta_g$ following best practices from~\cite{fjord} using FedAvg and finetune it on the same set of clients. Table~\ref{tab:cifar} shows the personalized accuracy of the various fine-tuning baselines (Section~\ref{subsec:baseline}) and \method{} using $e=5 \& 15$ local epochs on groups of clients with varying label distribution, $P_i(y)$; $\alpha=1000$ represents the IID case and $\alpha=1.0,0.5,0.1$ represents more heterogeneous case. As observed in many previous works~\cite{fedbabu,jiang2019improving}, increasing label heterogeneity would result in a better initial global model at a expense of personalized performance. Our method instead retains the initial global performance and focuses on improving personalized performance.

\begin{figure}[h]
\centering
    
\begin{subfigure}{0.45\columnwidth}
    \includegraphics[trim=0 0 0 0, clip, width=0.97\columnwidth]{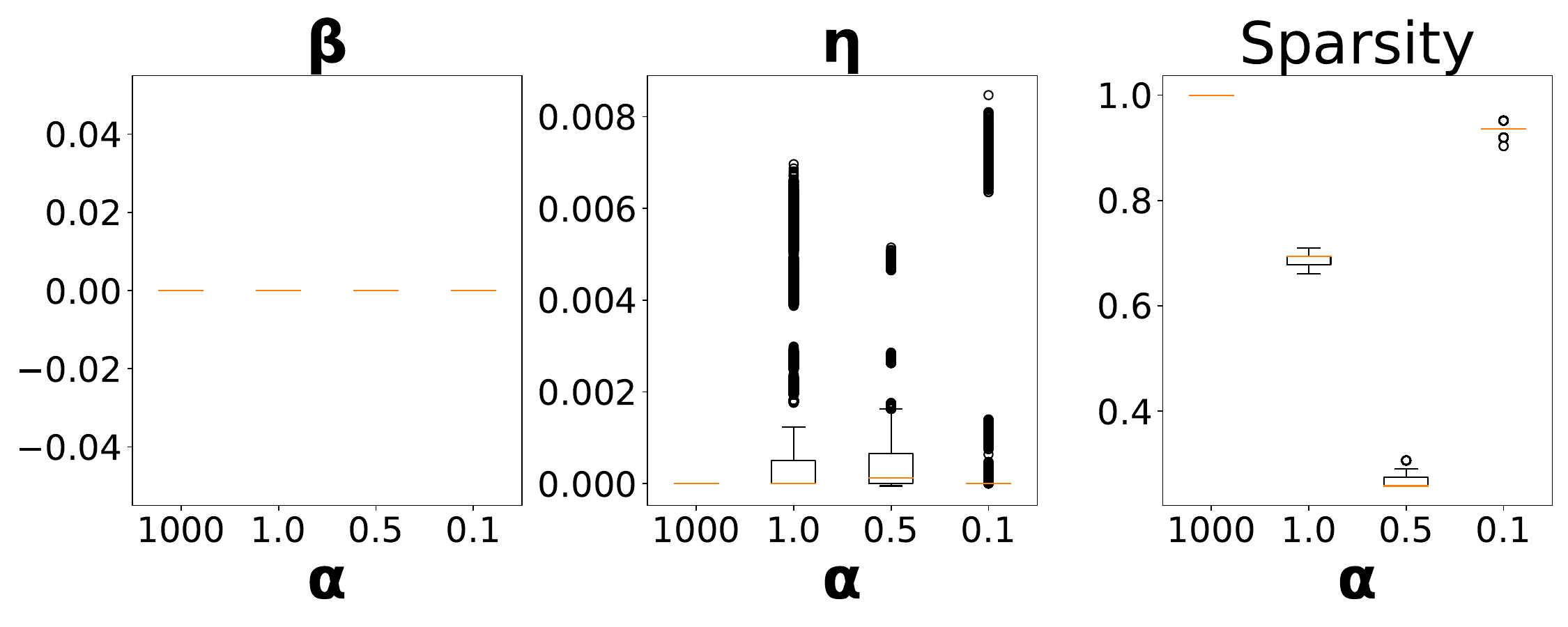}
    
    \vspace{-5pt}
    \caption{CIFAR10 ($e=15$)}
    \label{fig:hp_cifar}
\end{subfigure}
\vspace{-1pt}
\begin{subfigure}{0.45\columnwidth}
    \includegraphics[trim=0 0 0 0, clip, width=0.97\columnwidth]{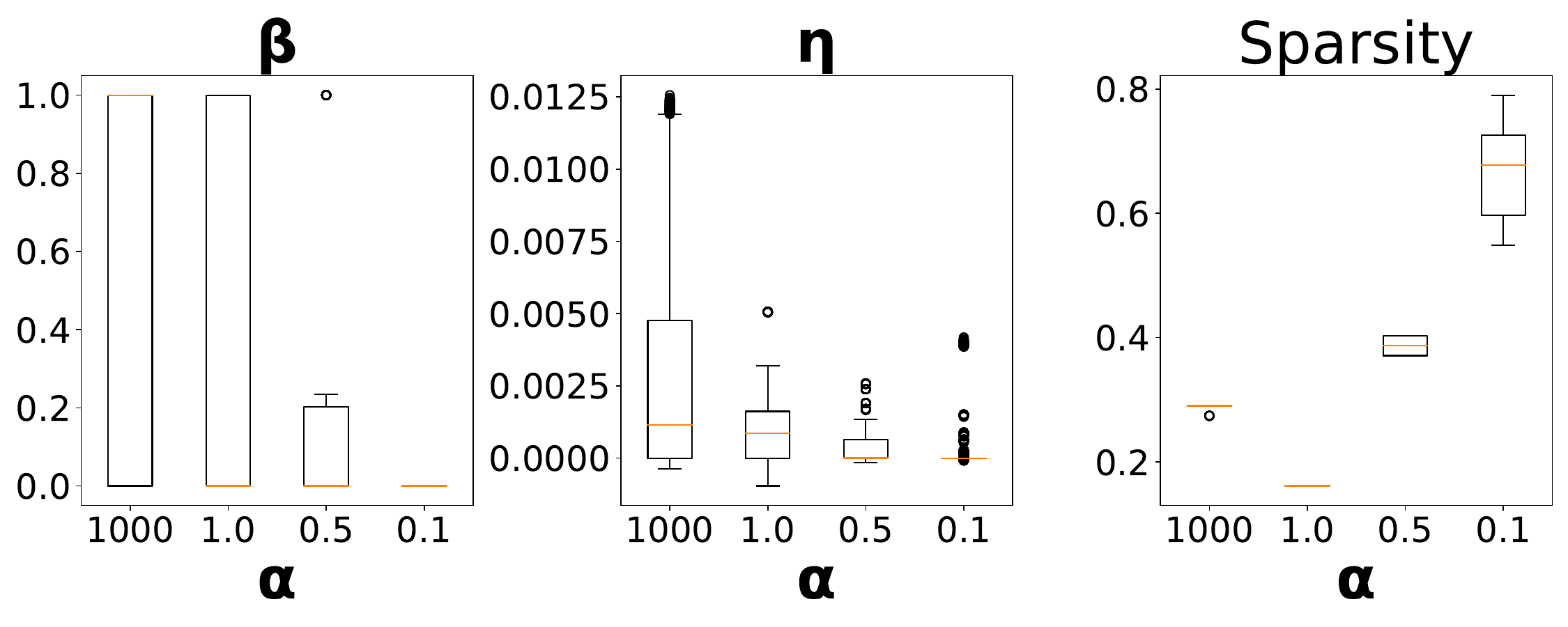} 
    
    \vspace{-5pt}
    \caption{CIFAR10C}
    \label{fig:hp_cifarC}
\end{subfigure}
\begin{subfigure}{0.45\columnwidth}
    \includegraphics[trim=0 0pt 0 0pt, clip,width=0.97\columnwidth]{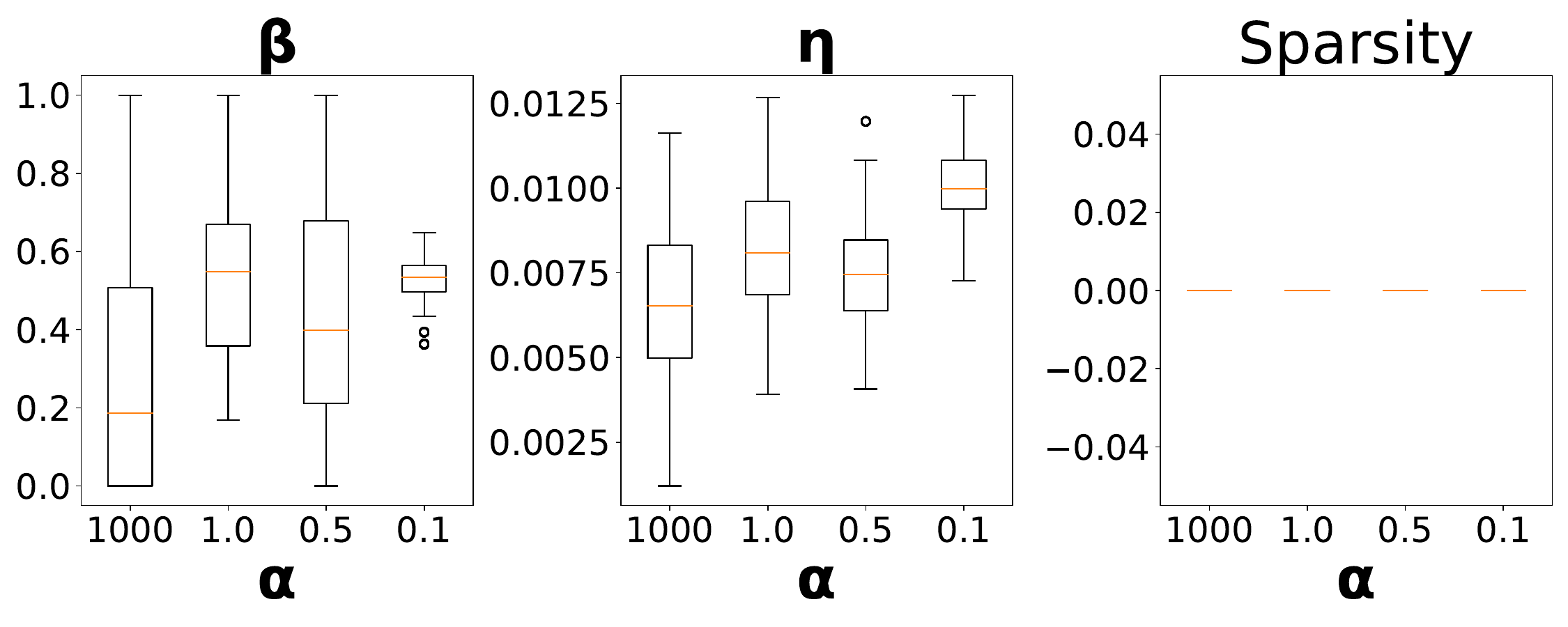}
    
    \vspace{-5pt}
    \caption{DomainNet}
    \label{fig:hp_domainnet}
\end{subfigure}
\begin{subfigure}{0.45\columnwidth}
    \includegraphics[trim=0 0pt 0 0pt, clip,width=\columnwidth]{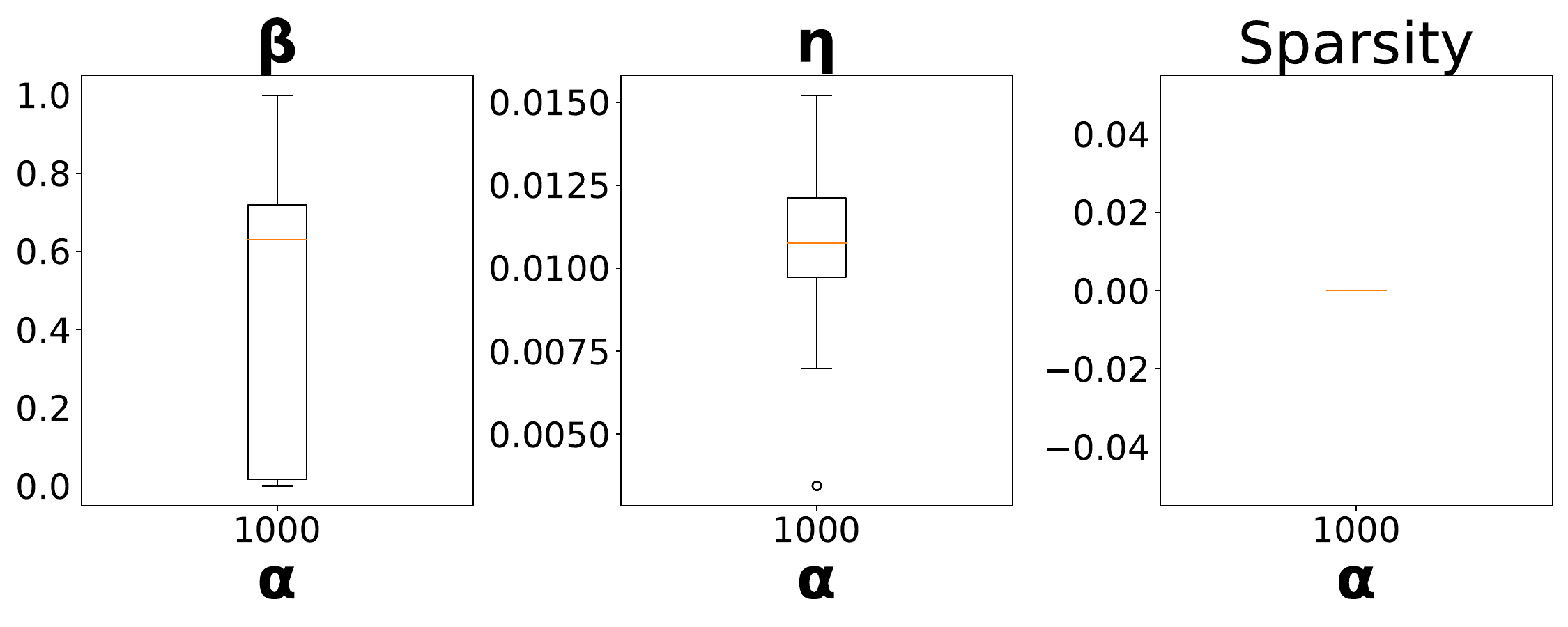}
    
    \vspace{-5pt}
    \caption{Office-Caltech-10}
    \label{fig:hp_office}
\end{subfigure}

\vspace{-0.5em}
\caption{Locality, spread, and skewness of \method{}'s learned hyperparameters, $\boldsymbol{\beta}$ and $\boldsymbol{\eta}$, of different layers across clients and the model's sparsity of all clients for each personalization scenario.}
\label{fig:hp}
\vspace{-1em}
\end{figure}

We also show that in many cases, especially for clients with limited local compute budget $e=5$, utilizing the pretrained model's BN statistics result (\textbf{BN G}) can be more beneficial as CIFAR-10 consists of images from the same natural image domain; in contrast, previous works mainly use either the client's BN statistics (\textbf{BN C}) or the incoming feature batch statistics (\textbf{BN I}) to normalize the features. This strategy is discovered by \method{}, as illustrated in Fig.~\ref{fig:hp_cifar} where the learned $\boldsymbol{\beta}$ is $0$ for all BN layers of all clients. For the IID case in particular, \method{} learns a sparsity\footnote{Sparsity refers to the percent of parameters whose learned learning rate for FT is 0.} of $1.0$, learning rate $\eta=0$, for all layers in all clients, forgoing fine-tuning and using the initial global model as the personalized model. For $\alpha=1.0$ and $0.1$, \method{} learns highly sparse models similar to recent works that proposed fine-tuning only a subset of hand-picked layers~\cite{fedbabu,ifca,fedrep,fedper,lgfedavg} to obtain performance gains.
Lastly, \methodnofed{} performs worse than some standard fine-tuning baselines as it meta-overfits on each client's validation set, highlighting the benefits of FL over local HPO. 

\textbf{\method{}'s Complementability with previous FL works.} As our proposed \method{} learns to improve the FT process, it is complementary, not competing, with other FL methods that learn shared model(s). Hence, besides FedAvg, we utilize \method{} to better personalize $\theta_g$ pretrained using PerFedAvg(HF)~\cite{perfedavg} and FedBABU~\cite{fedbabu} as shown in Table.~\ref{tab:extend_cifar}, where we compare \method{} against the most commonly used FT approach, \textbf{BN C}. 
Our results show that applying FedL2P to all three FL methods can lead to further gains, in most cases outperforming FT in each respective method.
This performance improvement can also bridge the performance gap between different methods. For instance, while FedAvg+FT has worse performance than FedBABU+FT in all cases, FedAvg+\method{} obtained comparable or better performance than FedBABU+FT for $\alpha=1000$ \& $0.1$. 

\vspace{-0.7mm}

\subsection{Personalizing to Unseen Clients}\label{subsec:cifarc}
\vspace{-0.5em}
\input{tables/domain.tex}
\input{tables/commands.tex}

\textbf{Unseen during Pretraining.} We evaluate the performance of \method{} on CIFAR-10-C starting from the pretrained model trained using FedAvg on CIFAR-10 in the IID setting $\alpha=1000$ (Section.~\ref{subsec:cifar}) as shown in Table.~\ref{tab:domain}. These new clients, whose data is partitioned from CIFAR-10-C, did not participate in the pretraining of the global model and, instead, only participate in the training of the \metanets{}. As added noise would impact the feature shift among clients, Fig.~\ref{fig:hp_cifarC} shows \method{} learns to use $\beta=1$ as compared to sole use of $\beta=0$ in the CIFAR-10 case (Fig.~\ref{fig:hp_cifar}); some clients benefit more by using its own BN statistics.  
Hence, the performance gap for non-IID cases between \textbf{BN C} and \textbf{BN G} is significantly smaller as compared to CIFAR-10 shown in Table.~\ref{tab:cifar}.

\textbf{Unseen during Learning of Meta-nets.} We evaluate the case where some clients are not included in the learning of the \metanets{} through our experiments on the Speech Commands dataset. We first pretrain the model using FedAvg and then train the \metanets{} using our \textit{seen} pool of clients. The learned \metanets{} are then used to fine-tune and evaluate both our \textit{seen} and \textit{unseen} pool of clients as shown in Table.~\ref{tab:commands}. Note that \methodnofed{} is not evaluated on the \textit{unseen} pool of clients as we only evaluate the meta-nets learned on the \textit{seen} pool of clients. We see that fine-tuning the pretrained model on each client’s train set resulted in a significant drop in test set performance. Moreover, as each client represents a different speaker, fine-tuning using the pretrained model’s BN statistics (\textbf{BN G}) fails entirely. \method{}, on the other hand, led to a significant performance boost on both the \textit{seen} and \textit{unseen} client pools, e.g. by utilizing a mix of pretrained model’s BN statistics and the client’s BN statistics shown in Appendix~\ref{app:add_results}.

\input{tables/ablation.tex}

\vspace{-1.5mm}
\subsection{Personalizing to Different Domains}\label{subsec:officedomainnet}
\vspace{-0.5em}

\input{tables/ari.tex}

We evaluate our method on Office-Caltech-10 and DomainNet datasets commonly used in domain generalization/adaptation, which exhibit both marginal and conditional feature shifts. Differing from the conventional FL setup, we adopted a pretrained model trained using ImageNet~\cite{imagenet} 
and attached a prediction head as our global model. 
Similar to CIFAR-10-C experiments, we compare our method with the baselines in Table.~\ref{tab:domain}. Unlike CIFAR-10/10-C, Office-Caltech-10 and DomainNet consist of images from distinct domains and adapting to these domain result in a model sparsity of 0 as shown in Fig.~\ref{fig:hp_domainnet} and Fig.~\ref{fig:hp_office}. Hence, a better personalization performance is achieved using the client's own local BN statistics (\textbf{BN C}) or the incoming batch statistics (\textbf{BN I}) than the pretrained model's (\textbf{BN G}). Lastly, each client's personalized model uses an extent of the pretrained natural image statistics as well as its own domain statistics, for both datasets.

\begin{figure}
\vspace{-2.1em}
\centering
\begin{subfigure}{0.49\textwidth}
    \includegraphics[trim=0 100pt 0 100pt, clip, width=\textwidth]{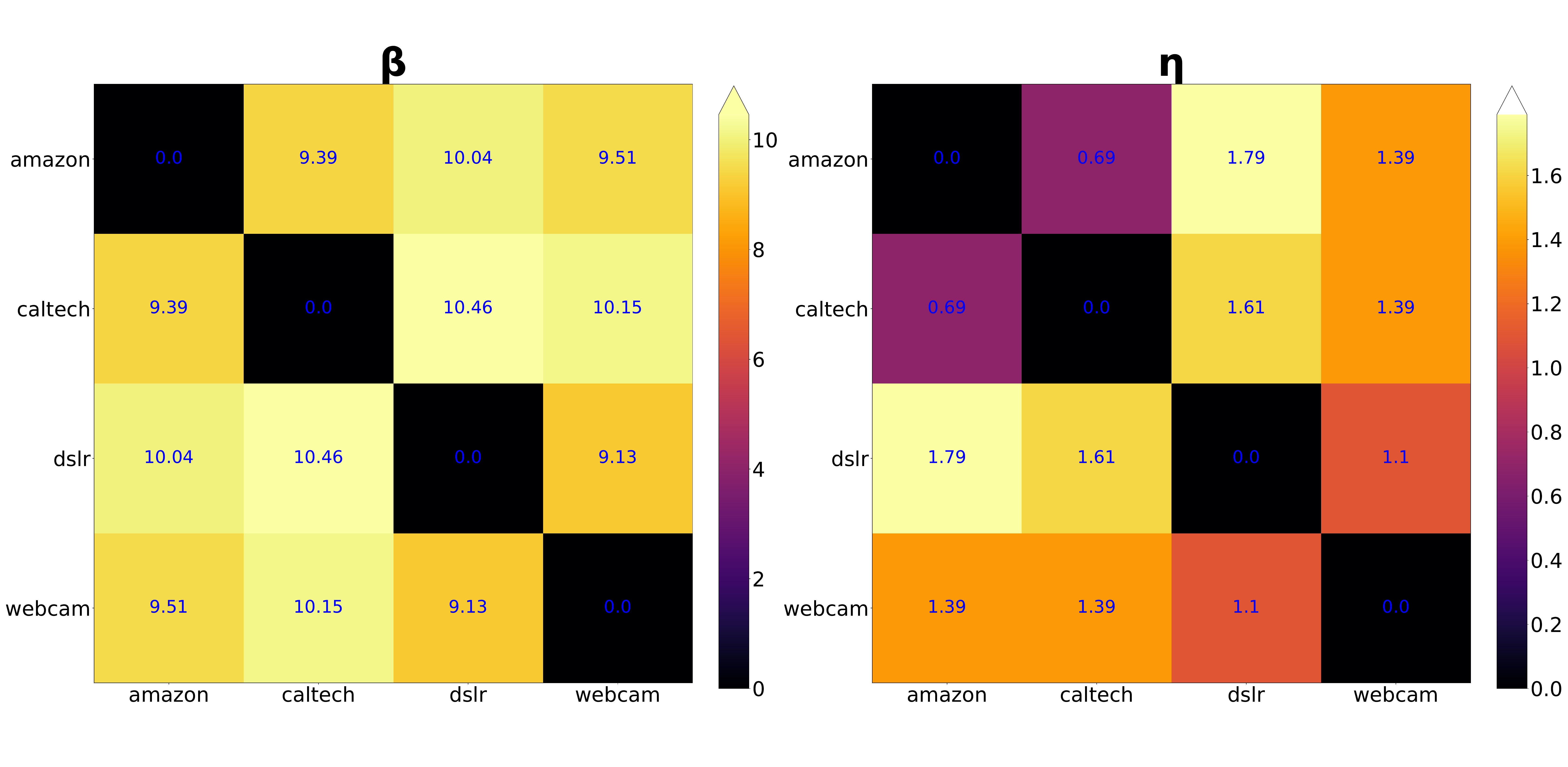}
    
    \vspace{-5pt}
    \caption{Office-Caltech-10}
    \label{fig:heatmap_office}
\end{subfigure}
\vspace{-5pt}
\begin{subfigure}{0.49\textwidth}
    \includegraphics[trim=0 100pt 0 100pt, clip,width=\textwidth]{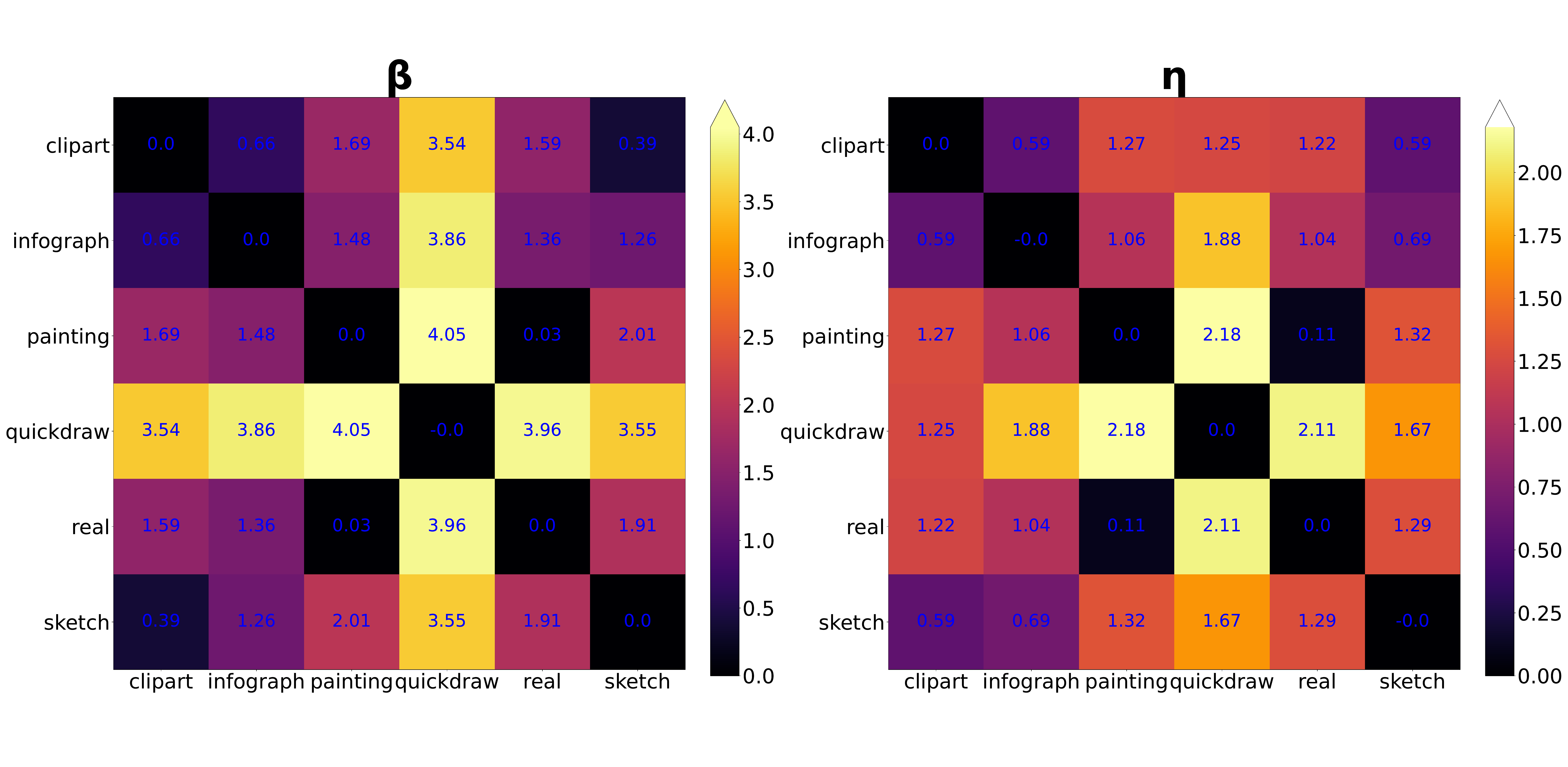}
    
    \vspace{-5pt}
    \caption{DomainNet ($\alpha=0.5$)}
    \label{fig:heatmap_domainnet0.5}
\end{subfigure}

\caption{
Cluster distance maps. 
Each block represents {\em normalized} distance, where the distance of the block $(j,k)$ is measured as the average of the Euclidean distances between all pairs of clients' $\boldsymbol{\beta}$ that belong to domain $j$ and domain $k$ (similarly for $\boldsymbol{\eta}$'s), normalized by the within-domain distance (see text). 
An off-diagonal block $> 0$ indicates that the corresponding clusters are better aligned with the true domains. 
}
\label{fig:heatmap}
\vspace{-1.5em}
\end{figure}

\textbf{Further Analysis}\quad 
To further investigate how well the hyperparameters returned by our \method{}-trained \metanets{} capture the domain-specific information, we analyze the clustering of local features, namely $\boldsymbol{\xi}=(\xi_1,\hdots,\xi_{B})$ and $\boldsymbol{x}=(E(x_0),SD(x_0),\hdots,E(x_{M-1}),SD(x_{M-1}))$ (i.e., the inputs to the \metanets{}), and the resulting respective hyperparameters, $\boldsymbol{\beta}$ and $\boldsymbol{\eta}$ (i.e., outputs of the \metanets{}), among clients. Specifically, we compute the similarity matrix using the Euclidean distance between the features/hyperparameters of any two clients and perform spectral clustering\footnote{
We use the default parameters for spectral clustering from \texttt{scikit-learn}~\cite{scikit-learn}.
}, using the discretization approach proposed in~\cite{stella2003multiclass} to assign labels on the normalized Laplacian embedding. We then compute the {\em Adjusted Rand Index} (ARI)~\cite{ari} between the estimated clusters and the clusters partitioned by the different data domains as shown in Table.~\ref{tab:ari}. 
We also visualize the alignment between the estimated clusters and the true domains in Fig.~\ref{fig:heatmap}, where each block $(j,k)$ represents the {\em normalized} average Euclidean distance between all pairs of clients in domain $j$ and $k$. Specifically, we divide the mean distance between domain $j$ and $k$ by the within-domain mean distances and take the log scale for better visualization: $\log(\frac{d(j,k)}{\sqrt{d(j,j)}\sqrt{d(k,k)}})$ where $d(j,k)$ is the mean Euclidean distance between $j$ and $k$. Thus, a random clustering has distance close to 0 and $>0$ indicates better alignment between the clustering and the true domains. 

As shown, for DomainNet, both BNNet and LRNet consistently preserve the cluster information found in their inputs, $\boldsymbol{\xi}$ \& $\boldsymbol{x}$, respectively. However, perfect clustering is not achieved due to the inherent difficulty. For instance, the \textit{real} and \textit{painting} domains share similar features, resulting in similar hyperparameters; the cross-domain distance between \textit{real} and \textit{painting} is $\sim0$ in log-scale in Fig.~\ref{fig:heatmap} and hence indistinguishable from their true domains. In contrast, the clients' features and resulting hyperparameters of the Office-Caltech-10 dataset are perfectly clustered (ARI=1) as visualized in Fig.~\ref{fig:heatmap_office} and Appendix.~\ref{app:add_results}.

\vspace{-0.5em}
\subsection{Ablation Study}\label{subsec:ablation}
\vspace{-0.5em}

To elucidate the individual impact of BNNet \& LRNet, we run an ablation study of all of the datasets used in our experiments and present the results in Table.~\ref{tab:ablation}, where CIFAR10 adopts the pretrained model trained using FedAvg. As BNNet learns to weight between client's BN statistics (\textbf{BN C}) and pretrained model's BN statistics (\textbf{BN G)}, running \method{} with BNNet alone leads to either better or similar performance to the better performing baseline. Running LRNet as a standalone, on the other hand, can result in further gains, surpassing the use of both BNNet and LRNet on some benchmarks. Nonetheless, it requires prior knowledge of the data feature distribution of the client in order to set a suitable $\boldsymbol{\beta}$, of which $\boldsymbol{\beta}=1$ uses \textbf{BN C} and $\boldsymbol{\beta}=0$ uses \textbf{BN G}. Our approach assumes no knowledge of the client's data and learns an estimated $\boldsymbol{\beta}$ per-scenario and per-client using BNNet.

%% file: tables/cifar_extended.tex
\begin{wraptable}{r}{0.45\textwidth}
\vspace{-4.5em}
\centering
\caption{\method{} complements existing FL methods by improving on the finetuning process. Experiments on CIFAR10 ($e=15$).}
\vspace{-0.5em}
\begin{scriptsize}
\resizebox{0.86\textwidth}{!}{\begin{minipage}{\textwidth}
\begin{tabular}{|cccc|}
\hline
  \multicolumn{1}{|c|}{\textbf{$\boldsymbol{\alpha}$}} &
\multicolumn{1}{|c|}{\textbf{Approach}} & \multicolumn{1}{c|}{\textbf{+FT (BN C)}} & \textbf{+\method{}} \\ \hline
\multicolumn{1}{|c|}{\textbf{1000}} &
\multicolumn{1}{|c|}{FedAvg} & \multicolumn{1}{c|}{63.04±0.02} & 65.13±0.02 \\ 
\multicolumn{1}{|c|}{\tiny{($\downarrow$ heterogeneity)}} &
\multicolumn{1}{|c|}{PerFedAvg(HF)} & \multicolumn{1}{c|}{34.58±0.13} & 47.58±0.01 \\ 
\multicolumn{1}{|c|}{} &
\multicolumn{1}{|c|}{FedBABU} & \multicolumn{1}{c|}{65.00±0.07} & 66.49±0.03 \\ \hline

\multicolumn{1}{|c|}{\textbf{1.0}} &
\multicolumn{1}{|c|}{FedAvg} & \multicolumn{1}{c|}{61.42±0.13} & 65.76±0.31 \\ 
\multicolumn{1}{|c|}{} &
\multicolumn{1}{|c|}{PerFedAvg(HF)} & \multicolumn{1}{c|}{44.85±0.28} & 50.2±1.26 \\ 
\multicolumn{1}{|c|}{} &
\multicolumn{1}{|c|}{FedBABU} & \multicolumn{1}{c|}{68.92±0.11} & 70.71±0.28 \\ \hline
\multicolumn{1}{|c|}{\textbf{0.5}} &
\multicolumn{1}{|c|}{FedAvg} & \multicolumn{1}{c|}{62.34±0.14} & 68.45±0.5 \\ 
\multicolumn{1}{|c|}{} &
\multicolumn{1}{|c|}{PerFedAvg(HF)} & \multicolumn{1}{c|}{52.43±0.16} & 55.05±0.53 \\ 
\multicolumn{1}{|c|}{} &
\multicolumn{1}{|c|}{FedBABU} & \multicolumn{1}{c|}{72.26±0.1} & 72.87±0.42 \\ \hline
\multicolumn{1}{|c|}{\textbf{0.1}} &
\multicolumn{1}{|c|}{FedAvg} & \multicolumn{1}{c|}{79.15±0.07} & 80.28±0.07 \\ 
\multicolumn{1}{|c|}{\tiny{($\uparrow$ heterogeneity)}} &
\multicolumn{1}{|c|}{PerFedAvg(HF)} & \multicolumn{1}{c|}{77.31±0.05} & 77.68±0.13 \\ 
\multicolumn{1}{|c|}{} &
\multicolumn{1}{|c|}{FedBABU} & \multicolumn{1}{c|}{79.50±0.08} & 79.58±0.04 \\ \hline
\end{tabular}
\end{minipage}
}
\end{scriptsize}
\label{tab:extend_cifar}
\vspace{-1.0em}
\end{wraptable}

%% file: tables/cifar.tex
\begin{table}[!t]
\vspace{-3.5em}

\caption{Experiments on CIFAR-10 using pretrained model trained using FedAvg~\cite{fedavg}. Both initial \& personalized accuracies are learnt and personalized on the same set of clients.}
\label{tab:cifar}
\vspace{0.2em}
\begin{scriptsize}
\resizebox{1.0\textwidth}{!}{\begin{minipage}{\textwidth}
\begin{center}
\begin{tabular}{|cccccccc|}
\hline
  \multicolumn{1}{|c|}{\textbf{$\boldsymbol{\alpha}$}} &
\multicolumn{1}{c|}{\textbf{\begin{tabular}[c]{@{}c@{}}Epochs (e)\end{tabular}}} &
  \multicolumn{1}{c|}{\textbf{\begin{tabular}[c]{@{}c@{}}Global Accuracy\end{tabular}}} &
  \multicolumn{1}{c|}{\textbf{+FT (BN C)}} &
  \multicolumn{1}{c|}{\textbf{+FT (BN G)}} &
  \multicolumn{1}{c|}{\textbf{+FT (BN I)}} &
  \multicolumn{1}{c|}{\textbf{+\methodnofed}} &
  \textbf{+\method} \\ \hline
\multicolumn{1}{|c|}{\textbf{1000}} &
\multicolumn{1}{c|}{5} &
  \multicolumn{1}{c|}{65.13} &
  \multicolumn{1}{c|}{64.35±0.03} &
  \multicolumn{1}{c|}{62.14±0.13} &
  \multicolumn{1}{c|}{56.58±0.08} &
  \multicolumn{1}{c|}{53.61±0.12} &
  \multicolumn{1}{l|}{\textbf{64.53±0.06}} \\

\multicolumn{1}{|c|}{\tiny{($\downarrow$ heterogeneity)}} &
\multicolumn{1}{c|}{15} &
  \multicolumn{1}{c|}{65.13} &
  \multicolumn{1}{c|}{63.04±0.02} &
  \multicolumn{1}{c|}{59.85±0.04} &
  \multicolumn{1}{c|}{55.72±0.03} &
  \multicolumn{1}{c|}{58.41±0.38} &
  \multicolumn{1}{l|}{\textbf{65.13±0.02}} \\ \hline
\multicolumn{1}{|c|}{\textbf{1.0}} &
\multicolumn{1}{c|}{5} &
  \multicolumn{1}{c|}{60.19} &
  \multicolumn{1}{c|}{59.72±0.18} &
  \multicolumn{1}{c|}{63.45±0.04} &
  \multicolumn{1}{c|}{50.8±0.06} &
  \multicolumn{1}{c|}{55.81±0.03} &
  \textbf{66.05±0.09} \\

\multicolumn{1}{|c|}{} &
\multicolumn{1}{c|}{15} &
  \multicolumn{1}{c|}{60.19} &
  \multicolumn{1}{c|}{61.42±0.13} &
  \multicolumn{1}{c|}{63.23±0.15} &
  \multicolumn{1}{c|}{54.94±0.07} &
  \multicolumn{1}{c|}{61.77±0.25} &
  \textbf{65.76±0.31} \\ \hline
\multicolumn{1}{|c|}{\textbf{0.5}} &
\multicolumn{1}{c|}{5} &
  \multicolumn{1}{c|}{57.12} &
  \multicolumn{1}{c|}{58.22±0.02} &
  \multicolumn{1}{c|}{67.16±0.11} &
  \multicolumn{1}{c|}{50.33±0.01} &
  \multicolumn{1}{c|}{59.79±0.09} &
  \textbf{68.96±0.09} \\
\multicolumn{1}{|c|}{} &
\multicolumn{1}{c|}{15} &
  \multicolumn{1}{c|}{57.12} &
  \multicolumn{1}{c|}{62.34±0.14} &
  \multicolumn{1}{c|}{67.4±0.06} &
  \multicolumn{1}{c|}{58.12±0.07} &
  \multicolumn{1}{c|}{65.28±0.23} &
  \textbf{68.45±0.5} \\ \hline
\multicolumn{1}{|c|}{\textbf{0.1}} &
\multicolumn{1}{c|}{5} &
  \multicolumn{1}{c|}{44.86} &
  \multicolumn{1}{c|}{68.04±0.05} &
  \multicolumn{1}{c|}{78.73±0.04} &
  \multicolumn{1}{c|}{61.91±0.06} &
  \multicolumn{1}{c|}{74.26±0.29} &
  \textbf{80.33±0.13} \\
\multicolumn{1}{|c|}{\tiny{($\uparrow$ heterogeneity)}} &
\multicolumn{1}{c|}{15} &
  \multicolumn{1}{c|}{44.86} &
  \multicolumn{1}{c|}{79.15±0.07} &
  \multicolumn{1}{c|}{78.97±0.07} &
  \multicolumn{1}{c|}{75.94±0.0} &
  \multicolumn{1}{c|}{78.6±0.21} &
  \textbf{80.28±0.07} \\ \hline
\end{tabular}
\end{center}
\end{minipage}
}
\end{scriptsize}
\vspace{-1.5em}
\end{table}

%% file: tables/domain.tex
\begin{table}[!t]
\vspace{-3.5em}
\centering
\caption{Personalized test accuracies of CIFAR-10-C, Office-Caltech-10, Domainnet ($e=15$).}
\label{tab:domain}
\vspace{0.2em}
\begin{scriptsize}
\resizebox{1.0\textwidth}{!}{\begin{minipage}{\textwidth}
\begin{center}
\begin{tabular}{|ccccccc|}
\hline

\multicolumn{1}{|c|}{\textbf{$\boldsymbol{\alpha}$}} &
\multicolumn{1}{|c|}{\textbf{Dataset}} &
  \multicolumn{1}{c|}{\textbf{+FT (BN C)}} &
  \multicolumn{1}{c|}{\textbf{+FT (BN G)}} &
  \multicolumn{1}{c|}{\textbf{+FT (BN I)}} &
  \multicolumn{1}{c|}{\textbf{+\methodnofed}} &
  \textbf{+\method} \\ \hline
\multicolumn{1}{|c|}{\textbf{1000}} &
\multicolumn{1}{|c|}{CIFAR-10-C} &
  \multicolumn{1}{c|}{59.58±0.03} &
  \multicolumn{1}{c|}{57.03±0.08} &
  \multicolumn{1}{c|}{55.77±0.12} &
  \multicolumn{1}{c|}{58.80±0.13} &
  \multicolumn{1}{l|}{\textbf{59.97±0.22}} \\
  \multicolumn{1}{|c|}{\small{($\downarrow$ heterogeneity)}} &
\multicolumn{1}{|c|}{Caltech-10} &
  \multicolumn{1}{c|}{80.97±0.33} &
  \multicolumn{1}{c|}{36.02±25.21} &
  \multicolumn{1}{c|}{81.43±2.16} &
  \multicolumn{1}{c|}{75.52±3.83} &
  \multicolumn{1}{l|}{\textbf{88.85±0.89}} \\
  \multicolumn{1}{|c|}{} &
\multicolumn{1}{|l|}{DomainNet} &
  \multicolumn{1}{c|}{52.17±1.55} &
  \multicolumn{1}{c|}{30.55±1.07} &
  \multicolumn{1}{c|}{50.47±0.88} &
  \multicolumn{1}{c|}{45.04±1.56} &
  \multicolumn{1}{l|}{\textbf{54.38±0.45}} \\ \hline
\multicolumn{1}{|c|}{\textbf{1.0}} &
\multicolumn{1}{|c|}{CIFAR-10-C} &
  \multicolumn{1}{c|}{67.37±0.08} &
  \multicolumn{1}{c|}{66.45±0.03} &
  \multicolumn{1}{c|}{63.56±0.07} &
  \multicolumn{1}{c|}{66.63±0.09} &
  \textbf{68.83±0.15} \\
  \multicolumn{1}{|c|}{} &
\multicolumn{1}{|c|}{DomainNet} &
  \multicolumn{1}{c|}{62.27±0.58} &
  \multicolumn{1}{c|}{44.15±0.11} &
  \multicolumn{1}{c|}{59.4±0.7} &
  \multicolumn{1}{c|}{54.75±0.12} &
  \textbf{63.77±0.44} \\ \hline
\multicolumn{1}{|c|}{\textbf{0.5}} &
\multicolumn{1}{|c|}{CIFAR-10-C} &
  \multicolumn{1}{c|}{74.92±0.08} &
  \multicolumn{1}{c|}{75.24±0.17} &
  \multicolumn{1}{c|}{71.38±0.01} &
  \multicolumn{1}{c|}{74.48±0.06} &
  \textbf{76.78±0.22} \\
  \multicolumn{1}{|c|}{} &
\multicolumn{1}{|c|}{DomainNet} &
  \multicolumn{1}{c|}{71.39±0.97} &
  \multicolumn{1}{c|}{49.81±1.98} &
  \multicolumn{1}{c|}{68.94±0.71} &
  \multicolumn{1}{c|}{66.38±0.78} &
  \textbf{72.64±0.3} \\ \hline
\multicolumn{1}{|c|}{\textbf{0.1}} &
\multicolumn{1}{|c|}{CIFAR-10-C} &
  \multicolumn{1}{c|}{87.25±0.06} &
  \multicolumn{1}{c|}{88.5±0.02} &
  \multicolumn{1}{c|}{83.93±0.04} &
  \multicolumn{1}{c|}{87.93±0.31} &
  \textbf{89.23±0.15} \\
  \multicolumn{1}{|c|}{\small{($\uparrow$ heterogeneity)}} &
\multicolumn{1}{|c|}{DomainNet} &
  \multicolumn{1}{c|}{86.03±0.47} &
  \multicolumn{1}{c|}{69.41±1.95} &
  \multicolumn{1}{c|}{85.35±1.14} &
  \multicolumn{1}{c|}{83.93±1.02} &
  \textbf{86.36±0.45} \\ \hline
\end{tabular}
\end{center}
\end{minipage}
}
\end{scriptsize}
\vspace{-2.0em}
\end{table}

%% file: tables/commands.tex
\begin{table}[h]

\centering
\caption{Experiments on Speech Commands. Both the pretrained model and meta-nets are only learned on the \textit{seen} pool of clients.}
\label{tab:commands}
\begin{scriptsize}
\resizebox{0.85\textwidth}{!}{\begin{minipage}{\textwidth}
\begin{center}
\begin{tabular}{|cccccccc|}
\hline
  \multicolumn{1}{|c|}{\textbf{Client Pool}} &
\multicolumn{1}{c|}{\textbf{\begin{tabular}[c]{@{}c@{}}Epochs (e)\end{tabular}}} &
  \multicolumn{1}{c|}{\textbf{\begin{tabular}[c]{@{}c@{}}Global Accuracy\end{tabular}}} &
  \multicolumn{1}{c|}{\textbf{+FT (BN C)}} &
  \multicolumn{1}{c|}{\textbf{+FT (BN G)}} &
  \multicolumn{1}{c|}{\textbf{+FT (BN I)}} &
  \multicolumn{1}{c|}{\textbf{+\methodnofed}} &
  \textbf{+\method} \\ \hline
\multicolumn{1}{|c|}{\textit{Seen}} &
\multicolumn{1}{c|}{5} &
  \multicolumn{1}{c|}{82.11} &
  \multicolumn{1}{c|}{70.07±0.63} &
  \multicolumn{1}{c|}{2.98±0.00} &
  \multicolumn{1}{c|}{65.97±0.43} &
  \multicolumn{1}{c|}{64.79±1.84} &
  \multicolumn{1}{l|}{\textbf{87.77±0.47}} \\

\multicolumn{1}{|c|}{} &
\multicolumn{1}{c|}{15} &
  \multicolumn{1}{c|}{82.11} &
  \multicolumn{1}{c|}{74.15±0.70} &
  \multicolumn{1}{c|}{2.98±0.0} &
  \multicolumn{1}{c|}{68.98±0.19} &
  \multicolumn{1}{c|}{65.80±1.66} &
  \multicolumn{1}{l|}{\textbf{84.74±0.38}} \\ \hline
\multicolumn{1}{|c|}{\textit{Unseen}} &
\multicolumn{1}{c|}{5} &
  \multicolumn{1}{c|}{81.62} &
  \multicolumn{1}{c|}{62.76±1.83} &
  \multicolumn{1}{c|}{2.85±0.00} &
  \multicolumn{1}{c|}{64.88±0.66} &
  \multicolumn{1}{c|}{-} &
  \textbf{87.85±0.18} \\

\multicolumn{1}{|c|}{} &
\multicolumn{1}{c|}{15} &
  \multicolumn{1}{c|}{81.62} &
  \multicolumn{1}{c|}{68.76±3.40} &
  \multicolumn{1}{c|}{2.85±0.00} &
  \multicolumn{1}{c|}{67.28±0.46} &
  \multicolumn{1}{c|}{-} &
  \textbf{84.60±0.35} \\ \hline
\end{tabular}
\end{center}
\end{minipage}
}
\end{scriptsize}
\vspace{-1.5em}
\end{table}

%% file: tables/ablation.tex
\begin{table}[h]
\vspace{-1.0em}
\caption{Ablation study 
for \method~with $e=15$.}
\centering
\label{tab:ablation}
\vspace{0.2em}
\begin{scriptsize}
\resizebox{0.85\textwidth}{!}{\begin{minipage}{\textwidth}
\begin{center}
\begin{tabular}{|cccccccc|}
\hline
  \multicolumn{1}{|c|}{\textbf{$\boldsymbol{\alpha}$}} &
\multicolumn{1}{|c|}{\textbf{Dataset}} &
  \multicolumn{1}{c|}{\textbf{+FT (BN C)}} &
  \multicolumn{1}{c|}{\textbf{+FT (BN G)}} &
  \multicolumn{1}{c|}{\textbf{\begin{tabular}[c]{@{}c@{}}+\method \\ (BNNet)\end{tabular}}} &
  \multicolumn{1}{c|}{\textbf{\begin{tabular}[c]{@{}c@{}}+\method\\ (LRNet)  $\boldsymbol{\beta}$=1\end{tabular}}} &
  \multicolumn{1}{c|}{\textbf{\begin{tabular}[c]{@{}c@{}}+\method\\ (LRNet) $\boldsymbol{\beta}$=0\end{tabular}}} &
  \textbf{+\method} \\ \hline
\multicolumn{1}{|c|}{\textbf{1000}} &
\multicolumn{1}{|c|}{CIFAR-10} &
  \multicolumn{1}{c|}{63.04±0.02} &
  \multicolumn{1}{c|}{59.85±0.04} &
  \multicolumn{1}{c|}{62.35±0.24} &
  \multicolumn{1}{c|}{62.62±0.21} &
  \multicolumn{1}{c|}{65.11±0.02} &
  \textbf{65.13±0.02} \\
\multicolumn{1}{|c|}{\tiny{($\downarrow$ heterogeneity)}} &
\multicolumn{1}{|c|}{CIFAR-10-C} &
  \multicolumn{1}{c|}{59.58±0.03} &
  \multicolumn{1}{c|}{57.03±0.08} &
  \multicolumn{1}{c|}{59.57±0.13} &
  \multicolumn{1}{c|}{\textbf{60.09±0.02}} &
  \multicolumn{1}{c|}{59.30±0.11} &
  59.97±0.22 \\
  \multicolumn{1}{|c|}{} &
\multicolumn{1}{|c|}{Caltech-10} &
  \multicolumn{1}{c|}{80.97±0.33} &
  \multicolumn{1}{c|}{36.02±25.21} &
  \multicolumn{1}{c|}{88.12±1.18} &
  \multicolumn{1}{c|}{85.50±5.76} &
  \multicolumn{1}{c|}{42.59±22.87} &
  \textbf{88.85±0.89} \\
  \multicolumn{1}{|c|}{} &
\multicolumn{1}{|l|}{DomainNet} &
  \multicolumn{1}{c|}{52.17±1.55} &
  \multicolumn{1}{c|}{30.55±1.07} &
  \multicolumn{1}{c|}{53.39±0.85} &
  \multicolumn{1}{c|}{\textbf{55.59±2.76}} &
  \multicolumn{1}{c|}{44.43±3.46} &
  54.38±0.45 \\ \hline
\multicolumn{1}{|c|}{\textbf{1.0}} &
\multicolumn{1}{|c|}{CIFAR-10} &
  \multicolumn{1}{c|}{61.42±0.13} &
  \multicolumn{1}{c|}{63.23±0.15} &
  \multicolumn{1}{c|}{63.75±0.04} &
  \multicolumn{1}{c|}{64.67±0.06} &
  \multicolumn{1}{c|}{64.61±0.49} &
  \textbf{65.76±0.31} \\
    \multicolumn{1}{|c|}{} &
\multicolumn{1}{|c|}{CIFAR-10-C} &
  \multicolumn{1}{c|}{67.37±0.08} &
  \multicolumn{1}{c|}{66.45±0.03} &
  \multicolumn{1}{c|}{68.1±0.07} &
  \multicolumn{1}{c|}{68.62±0.07} &
  \multicolumn{1}{c|}{67.82±0.1} &
  \textbf{68.83±0.15} \\
    \multicolumn{1}{|c|}{} &
\multicolumn{1}{|c|}{DomainNet} &
  \multicolumn{1}{c|}{62.27±0.58} &
  \multicolumn{1}{c|}{44.15±0.11} &
  \multicolumn{1}{c|}{62.73±0.51} &
  \multicolumn{1}{c|}{63.69±0.43} &
  \multicolumn{1}{c|}{diverge} &
  \textbf{63.77±0.44} \\ \hline
\multicolumn{1}{|c|}{\textbf{0.5}} &
\multicolumn{1}{|c|}{CIFAR-10} &
  \multicolumn{1}{c|}{62.34±0.14} &
  \multicolumn{1}{c|}{67.4±0.06} &
  \multicolumn{1}{c|}{67.59±0.15} &
  \multicolumn{1}{c|}{\textbf{68.81±0.05}} &
  \multicolumn{1}{c|}{68.01±0.29} &
  68.45±0.50 \\
    \multicolumn{1}{|c|}{} &
\multicolumn{1}{|c|}{CIFAR-10-C} &
  \multicolumn{1}{c|}{74.92±0.08} &
  \multicolumn{1}{c|}{75.24±0.17} &
  \multicolumn{1}{c|}{76.36±0.08} &
  \multicolumn{1}{c|}{\textbf{76.86±0.06}} &
  \multicolumn{1}{c|}{76.11±0.07} &
  76.82±0.19 \\
    \multicolumn{1}{|c|}{} &
\multicolumn{1}{|c|}{DomainNet} &
  \multicolumn{1}{c|}{71.39±0.97} &
  \multicolumn{1}{c|}{49.81±1.98} &
  \multicolumn{1}{c|}{70.99±1.15} &
  \multicolumn{1}{c|}{\textbf{72.74±0.51}} &
  \multicolumn{1}{c|}{diverge} &
  72.64±0.30 \\ \hline
\multicolumn{1}{|c|}{\textbf{0.1}} &
\multicolumn{1}{|c|}{CIFAR-10} &
  \multicolumn{1}{c|}{79.15±0.07} &
  \multicolumn{1}{c|}{78.97±0.07} &
  \multicolumn{1}{c|}{79.47±0.2} &
  \multicolumn{1}{c|}{80.24±0.09} &
  \multicolumn{1}{c|}{\textbf{80.39±0.15}} &
  80.28±0.07 \\
\multicolumn{1}{|c|}{\tiny{($\uparrow$ heterogeneity)}} &
\multicolumn{1}{|c|}{CIFAR-10-C} &
  \multicolumn{1}{c|}{87.25±0.06} &
  \multicolumn{1}{c|}{88.5±0.02} &
  \multicolumn{1}{c|}{88.6±0.1} &
  \multicolumn{1}{c|}{89.08±0.04} &
  \multicolumn{1}{c|}{89.14±0.13} &
  \textbf{89.23±0.15} \\
    \multicolumn{1}{|c|}{} &
\multicolumn{1}{|c|}{DomainNet} &
  \multicolumn{1}{c|}{86.03±0.47} &
  \multicolumn{1}{c|}{69.41±1.95} &
  \multicolumn{1}{c|}{85.87±1.31} &
  \multicolumn{1}{c|}{85.78±0.6} &
  \multicolumn{1}{c|}{diverge} &
  \textbf{86.36±0.45} \\ \hline
\end{tabular}
\end{center}
\end{minipage}
}
\end{scriptsize}
\vspace{-0.8em}
\end{table}

%% file: tables/ari.tex


\begin{wraptable}{r}{0.45\textwidth}
\vspace{-2.9em}
\caption{ARI values between the clustering of clients by their domains and inputs/outputs of the BNNet and LRNet. 
}
\vspace{-0.5em}
\centering
\label{tab:ari}
\resizebox{0.6\textwidth}{!}{\begin{minipage}{\textwidth}
\begin{tabular}{cccccc}

\toprule
\multirow{2}{*}{Dataset} & \multirow{2}{*}{$\alpha$} & \multicolumn{2}{c}{BNNet} & \multicolumn{2}{c}{LRNet} \\
\cmidrule(lr){3-4} \cmidrule(lr){5-6}
& & Input ($\boldsymbol{\xi}$) & Output  ($\boldsymbol{\beta}$) & Input ($\boldsymbol{x}$) & Output ($\boldsymbol{\eta}$) \\
\hline
Caltech-10\Tstrut & 1000 & 1.0  & 1.0  & 1.0  & 1.0 \\
\hline
\multirow{4}{*}{DomainNet}\Tstrut & 1000 & 0.65 & 0.64 & 0.76 & 0.77 \\
 & 1.0  & 0.70 & 0.59 & 0.77 & 0.74 \\
 & 0.5  & 0.52 & 0.53 & 0.72 & 0.63 \\
 & 0.1  & 0.27 & 0.52 & 0.65 & 0.59 \\
\bottomrule
\end{tabular}
\end{minipage}}
\vspace{-1.0em}
\end{wraptable}

%% file: 05conclusion.tex
\vspace{-2mm}
\section{Conclusion}\label{sec:05conclusion}
\vspace{-1.0em}
In this paper, we propose \method, a framework for federated learning of personalization strategies specific to individual FL scenarios and datasets as well as individual clients. 
We learned \metanets{} that use clients' local data statistics relative to the pretrained model, to generate hyperparameters that explicitly target the normalization, scaling, and shifting of features as well as layer-wise parameter selection to mitigate the detrimental impacts of both marginal and conditional feature shift and marginal label shift, significantly boosting personalized performance. This framework is complementary to existing FL works that learn shared model(s) and can discover many previous hand-designed heuristics for sparse layer updates and BN parameter selection as special cases, and learns to apply them where appropriate according to the specific scenario for each specific client. As a future work, our approach can be extended to include other hyperparameters and model other forms of heterogeneity, e.g. using the number of samples as an expert input feature to a \metanet{}.

%% file: appendix/app_ift.tex
\section{Derivation of Equation (\ref{eq:ift})}
\label{app:ift}
From Eq (\ref{eq:obj}), we know that: 
\begin{align}\label{eq:app}
\frac{d \mathcal{L}_{i,T}(\theta_i^*,\lambda)}{d \theta} = 0
\end{align}

Based on the implicit functional theorem (IFT), we get that if we have a function $F(x,y) = c$, we can derive that $y'(x) = - F_x/ F_y $. Therefore, plus the Eq (\ref{eq:app}) into the theorem, we can get:

\begin{align}
\begin{split}
\frac{d\theta^*}{d \lambda} = - \frac{\partial_\theta ( \frac{d \mathcal{L}_{i,T}(\theta_i^*,\lambda)}{d \theta} )}{\partial_\lambda \frac{d \mathcal{L}_{i,T}(\theta_i^*,\lambda)}{d \theta}  } = -(\partial_\theta^2\mathcal{L}_{T}(\theta,\lambda))^{-1} \: \partial_{\lambda\theta} \mathcal{L}_{T}(\theta,\lambda)
\end{split}
\end{align}

%% file: appendix/app_usecase.tex
\section{Positioning of \method{}.}\label{app:usecase}

\input{tables/positioning.tex}

Table~\ref{tab:positioning} shows the positioning of \method{} against existing literature. Note that this list is by no means exhaustive but representative to highlight the position of our work. Most existing approaches obtained personalized models using a personalized policy and local data, often through a finetuning-based approach. This personalized policy can either 1)~be fixed, \textit{e.g.} hand-crafting hyperparameters, layers to freeze, selecting number of mixture components, number of clusters or 2)~learned, \textit{e.g.} learning a hypernetwork to generate weights or \metanets{} to generate hyperparameters. These approaches are also grouped based on whether this personalized policy is dependent on the local data during inference, \textit{e.g.} \metanets{} require local client meta-data to generate hyperparameters. 

In order to adapt to per-dataset per-client scenarios, many works rely on storing per-client personalized layers, which are trained only on each client's local data. Unfortunately, the memory cost of storing these models scales with the number of clients, C, restricting previous works to small scale experiments. We show that these works are impractical in our CIFAR-10 setup of 1000 clients in Appendix.~\ref{app:cost}. 
Moreover, most existing methods rely on a fixed personalized policy, such as deriving shared global hyperparameters for all clients in FLoRA, or they do not dependent on local data, such as FedEx which randomly samples per-client hyperparameters from learned categorical distributions. Hence, these methods do not adapt as well to per-dataset per-client scenarios and are ineffective at targeting both label and feature shifts. Lastly, although pFedHN and pFedLA target both label and feature shift cases, they do not scale well to our experiments as shown in Appendix.~\ref{app:cost}.

Most importantly, all existing FL approaches shown in Table~\ref{tab:positioning} can use finetuning either to personalize shared global model(s) or as a complementary personalization strategy to further adapt their personalized models. Since our proposed \method{} focus on better personalizing shared global model(s) by learning better a personalized policy which leverages the clients' local data statistics relative to the given pretrain model(s), our approach is complementary to all existing FL solutions that learn shared model(s); we showed improvements over a few of these works in Table~\ref{tab:extend_cifar}.

\textbf{Use Cases of \method{}}~Mainstream FL focuses on training from scratch, but we focus on federated learning of a strategy to adapt an existing pre-trained model (whether obtained by FL or not) on an unseen group of clients with heterogeneous data. There are several scenarios where this setup and our solution would be useful: 
\begin{enumerate}
\item Scenarios where it’s expensive to train from scratch for a new group of clients, e.g. adopting FedEx~\cite{fedex} from scratch for a new group of clients would require thousands of rounds to retrain the model and HP while our method takes hundreds to learn two tiny meta-nets (Appendix. \ref{app:cost}). 
\item Scenarios where there is a publicly available pre-trained foundation model that can be exploited. This is illustrated in Section~\ref{subsec:officedomainnet} where we adapt a publicly available pretrained model trained using ImageNet on domain generalization datasets. 
\item Scenarios where it’s important to also maintain a global model with high initial accuracy - often neglected by previous personalized FL works. 
\end{enumerate}

Note that our approach also does not critically depend on the global model's performance. Even in the worst case where the input statistics derived from the global model are junk (e.g., they degenerate to a constant vector, or are simply a random noise vector), then it just means the hyperparameters learned are no longer input-dependent. In this case, \method{} would effectively learn a constant vector of layer-wise learning rates + BN mixing ratio, as opposed to a function that predicts them. Thus, in this worst case we would lose the ability to customize the hyperparameters differently to different heterogeneous clients, but we would still be better off than the standard approach where these optimization hyperparameters are not learned. In the case where our global-model derived input features are better than this degenerate worst case, \method{}'s \metanets{} will improve on this already strong starting point.

%% file: tables/positioning.tex
\begin{table}[b!]
\caption{Positioning of \method{} with existing FL approaches. C is the total number of clients, M is the number of layers in the model, B is the number of BN layers in the model, D is the number of data domains, H is the number of hidden layers in the hypernetwork. }
\label{tab:positioning}
\vspace{0.2em}
\begin{scriptsize}
\resizebox{0.85\textwidth}{!}{\begin{minipage}{\textwidth}
\begin{center}
\begin{tabular}{|c|c|c|c|c|c|c|c|c|}
\hline
\textbf{\begin{tabular}[c]{@{}c@{}}FL \\ Approach\end{tabular}} & \textbf{\begin{tabular}[c]{@{}c@{}}Learns \\ Shared \\ Model(s)\end{tabular}} & \textbf{\begin{tabular}[c]{@{}c@{}}Personalized \\ Layers\end{tabular}} & \textbf{\begin{tabular}[c]{@{}c@{}}Personalized \\ Policy \\ Obtained by?\end{tabular}} & \textbf{\begin{tabular}[c]{@{}c@{}}Personalized \\ Policy \\ Data Dependent? \end{tabular}} & \textbf{\begin{tabular}[c]{@{}c@{}}Targets \\ Label Shift\end{tabular}} & \textbf{\begin{tabular}[c]{@{}c@{}}Targets \\ Feature Shift\end{tabular}} & \textbf{\begin{tabular}[c]{@{}c@{}}Memory \\ Cost\end{tabular}} & \textbf{\begin{tabular}[c]{@{}c@{}}Scale to \\ Large Networks\end{tabular}} \\ \hline
\begin{tabular}[c]{@{}c@{}}FedProx~\cite{fedprox}\\ PerFedAvg~\cite{perfedavg}\\ pFedMe~\cite{pfedme}\\ Ditto~\cite{ditto}\\ MOON~\cite{moon}\\ FedBABU~\cite{fedbabu} \end{tabular} & Yes & No & Fixed & No & Yes & No & O(M) & \cmark \\ \hline
PerFedMask~\cite{perfedmask} & Yes & No & Fixed & No & Yes & No & O(M) & \cmark \\ \hline
FedBN~\cite{fedbn} & Yes & Yes & Fixed & No & No & Yes & O(CB) & \cmark \\ \hline
\begin{tabular}[c]{@{}c@{}}FedPer~\cite{fedper}\\ FedRep~\cite{fedrep}\\ APFL~\cite{apfl}\\ LG-FedAvg~\cite{lgfedavg}\\ IFCA~\cite{ifca}\end{tabular} & Yes & Yes & Fixed & No & Yes & No & O(CM) & \xmark \\ \hline
FedDAR~\cite{feddar} & Yes & Yes & Fixed & No & Yes & Yes & O(DM) & \xmark \\ \hline
FLoRA~\cite{flora} & Yes & No & Fixed & No & Yes & No & O(M) & \cmark \\ \hline
FedEx~\cite{fedex} & Yes & Supported & Learned & No & Yes & No & O(M) & \cmark \\ \hline
FedEM~\cite{fedem} & Yes & No & Fixed & No & Yes & No & O(M) & \cmark \\ \hline
\begin{tabular}[c]{@{}c@{}}FedFOMO~\cite{fedfomo} \\ FedMe~\cite{fedme}\end{tabular} & No & Yes & Fixed & No & Yes & No & O(CM) & \xmark \\ \hline
\begin{tabular}[c]{@{}c@{}}pFedHN~\cite{pfedhn} \\ pFedLA~\cite{pFedLA}\end{tabular} & No & Yes & Learned & Yes & Yes & Yes & O(CMH) & \xmark \\ \hline

FedL2P (Ours) & No & Supported & Learned & Yes & Yes & Yes & O(M) & \cmark \\ \hline
\end{tabular}
\end{center}
\end{minipage}
}
\end{scriptsize}
\end{table}

%% file: appendix/app_cost.tex
\section{Cost of \method}\label{app:cost}

\textbf{Computational Cost of Hessian Approximation.}
We compare with hessian-free approaches, namely first-order (FO) MAML and hessian-free (HF) MAML, both of which are used by PerFedAvg, and measure the time it took to compute the meta-gradient after fine-tuning. Specifically, we run 100 iterations of each algorithm and report the mean of the walltime. Our proposed method takes 0.24 seconds to compute the hypergradient, 0.12 seconds of which is used to approximate the Hessian. In comparison, FO-MAML took 0.08 seconds and HF-MAML took 0.16 seconds to compute the meta-gradient. Hence, our proposed method is not a significant overhead relative to simpler non-Hessian methods. It is also worth noting that the number of fine-tuning epochs would not impact the cost of computing the hypergradient. 

\textbf{Memory Cost.}
In our CIFAR10 experiments, the meta-update of \method{} has a peak memory usage of 1.3GB. In contrast, existing FL methods that generate personalized policies require an order(s) of magnitude more memory and hence only evaluated in relatively small setups with smaller networks. For instance, pFedHN~\cite{pfedhn} requires in a peak memory usage of 17.93GB in our CIFAR10 setup as its user embeddings and hypernetwork scale up with the number of clients and model size. Moreover, they fail to generate reasonable client weights as these techniques do not scale to larger ResNets used in our experiments. APFL~\cite{apfl}, on the other hand, requires each client to maintain three models: local, global, and mixed personalized. Adopting APFL in our CIFAR10 setup of 1000 clients requires over 134GB of memory just to store the models per experiment, which is infeasible.

\textbf{Communication Cost.}
For each FL round, we transmit the parameters of the \metanets{}, which are lightweight MLP networks to from server to client and vice versa. Note that transmitting the global pretrained model to each new client is a one-time cost. Office-Caltech-10, DomainNet, and Speech Commands setups take a maximum of $100$ communication rounds, 0.24\% of the pretraining cost, to learn the \metanets{}. CIFAR-10 and CIFAR-10-C, on the other hand, can take hundreds of rounds up to a maximum of $500$ rounds, 0.38\% of the pretraining cost. In contrast, joint model and hyperparameter optimization typically takes thousands of rounds~\cite{fedex}, having to transmit both the model and the hyperparameter distribution across the network. 
In summary, \method{} incurs <1\% additional costs on top of pretraining and forgoes the cost of federatedly learning a model from scratch, which can be advantageous in certain scenarios as listed in Section.~\ref{app:usecase}.

\textbf{Inference Cost.}
During the fine-tuning stage, given the learned meta-nets, FedL2P requires 2 forward pass of the model per image and one forward pass of each meta-net to compute the personalized hyperparameters. This equates to ~0.55GFLOPs per image and would incur a minor additional cost of 4.4\% more than the regular finetuning process of 15 finetune epochs.

%% file: appendix/app_pretrain.tex
\section{Pretrained Model and Setup Details}\label{app:pretrain}

We use the Flower federated learning framework~\cite{flower} and 8 NVIDIA GeForce RTX 2080 Ti GPUs for all experiments. ResNet-18~\cite{resnet} is adopted with minor differences in the various setups:

\textbf{CIFAR-10.} We replaced the first convolution with a smaller convolution $3\times3$ kernel with stride$=1$ and padding$=1$ instead of the regular $7\times7$ kernel. We also replaced the max pooling operation with the identity operation and set the number of output features of the last fully connected layer to 10. The model is pretrained in a federated manner using FedAvg~\cite{fedavg} or FedBABU~\cite{fedbabu} or PerFedAvg(HF)~\cite{perfedavg} with a starting learning rate of $0.1$ for $500$ communication rounds. For PerFedAvg, we adopted the recommended hyperparameters used by the authors to meta-train the model. The fraction ratio is set to $r=0.1$; $100$ clients, each of who perform a single epoch update on its own local dataset before sending the updated model back to the server, participate per round. We dropped the learning rate by a factor of $0.1$ at round $250$ and $375$. This process is repeated for each $\alpha={1000,1.0,0.5,0.1}$, resulting in a pretrained model for each group of clients. We experiment with various fine-tuning learning rates $\{1.0,0.1,0.01,1e-3,1e-4,1e-5\}$ and pick the best-performing one, $1e-3$ for all experiments; the initial value of $\tilde{\boldsymbol{\eta}}$ in \method~is also set at $1e-3$.

\textbf{CIFAR-10-C.} We adopted the pretrained model trained in CIFAR-10 for $\alpha=1000$ and used the same fine-tuning learning rate for all experiments.

\textbf{Office-Caltech-10 \& DomainNet.} We adopted a Resnet-18 model that was pretrained on ImageNet~\cite{imagenet} and provided by torchvision~\cite{torchvision}. We replaced the number of output features of the last fully connected layer to 10. Similar to CIFAR-10 setup, we experiment with the same set of learning rates and pick the best-performing one, $1e-2$ for our experiments.

\textbf{Speech Commands.} We adopted the setup and hyperparameters from ZeroFL~\cite{zerofl}; a Resnet-18 model is trained using FedAvg for 500 rounds using a starting learning rate of $0.1$ and an exponential learning rate decay schedule with a base learning rate of $0.01$. We use the base learning rate for fine-tuning for all experiments.

%% file: appendix/app_arch.tex
\section{Architecture \& Initialization Details}\label{app:arch}

We present the architecture of our proposed \metanets, BNNet and LRNet. Both networks are 3-layer MLP models with 100 hidden layer neurons and ReLU~\cite{agarap2018deep} activations in-between layers. BNNet and LRNet clamp the output to a value of $[0,1]$ and $[0,1000]$ respectively and use a straight-through estimator~\cite{bengio2013estimating} (STE) to propagate gradients. We also tried using a sigmoid function for BNNet which converges to the same solution but at a much slower pace. We initialize the weights of BNNet and LRNet with Xavier initialization~\cite{glorot2010understanding} using the normal distribution with a gain value of $0.1$. To control the starting initial value of BNNet and LRNet, we initialize the biases of BNNet and LRNet with constants $0.5$ and $1.0$, resulting in initial values of $\sim0.5$ and $\sim1.0$ respectively. We also tried experimenting BNNet with different initializations by setting the biases to $[0.2, 0.5, 0.8]$ and got similar results.

%% file: appendix/app_results.tex
\section{Additional Results}\label{app:add_results}

\begin{figure*}[!t]
\centering
\begin{subfigure}{\textwidth}
    \includegraphics[trim=0 100pt 0 100pt, clip,width=\textwidth]{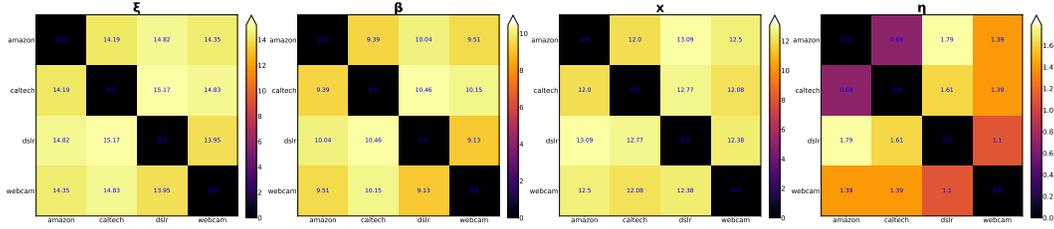}
    
    \caption{Office-Caltech-10 ($\alpha=1000$)}
\end{subfigure}
\begin{subfigure}{\textwidth}
    \includegraphics[trim=0 100pt 0 100pt, clip, width=\textwidth]{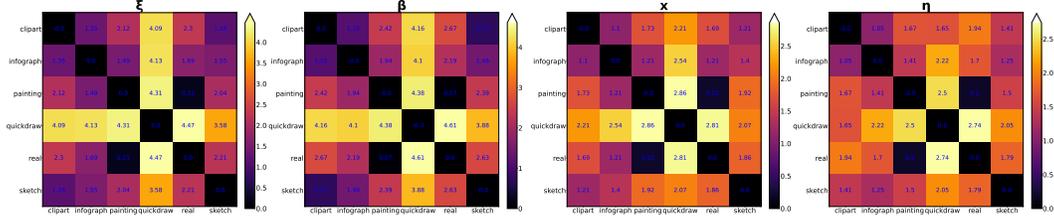}
    
    \caption{DomainNet ($\alpha=1000$)}
\end{subfigure}

\begin{subfigure}{\textwidth}
    \includegraphics[trim=0 100pt 0 100pt, clip,width=\textwidth]{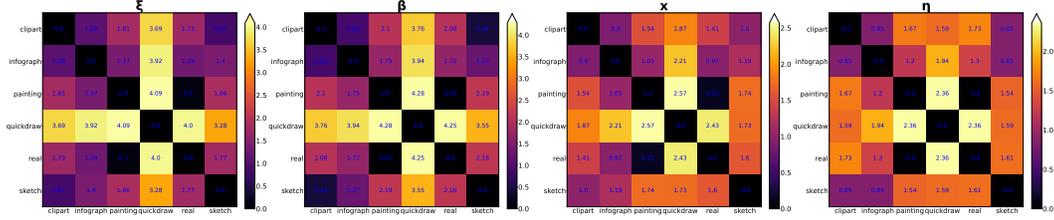}
    
    \caption{DomainNet ($\alpha=1.0$)}
\end{subfigure}
\begin{subfigure}{\textwidth}
    \includegraphics[trim=0 100pt 0 100pt, clip,width=\textwidth]{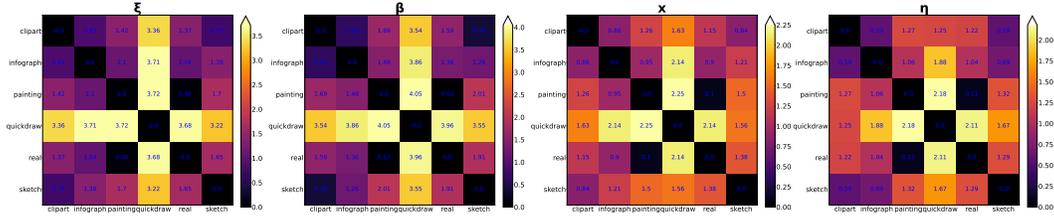}
    
    \caption{DomainNet ($\alpha=0.5$)}
\end{subfigure}
\begin{subfigure}{\textwidth}
    \includegraphics[trim=0 100pt 0 100pt, clip,width=\textwidth]{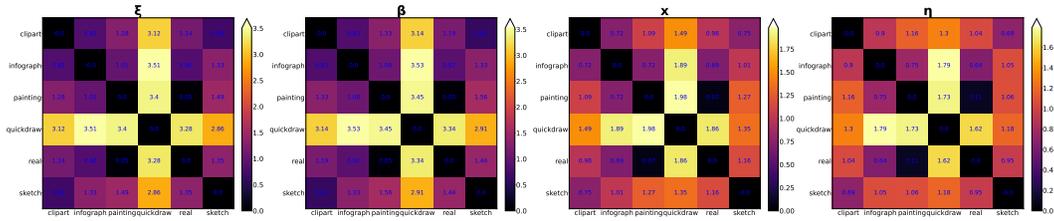}
    
    \caption{DomainNet ($\alpha=0.1$)}
\end{subfigure}
        
\caption{Cluster distance maps. Each block represents {\em normalized} distance between two domains (e.g., Caltech vs.~DSLR), where the distance of the block $(j,k)$ is measured as the average of the Euclidean distances between all pairs of clients' $\boldsymbol{\beta}$ that belong to domain $j$ and domain $k$ (similarly for $\boldsymbol{\eta}$'s). We normalize distance by the within-domain distances (see text), and take $\log$ for better visualization. Hence, an off-diagonal block greater than $0$ indicates that the corresponding clusters are better aligned with the true domains.}
\label{fig:app_heatmap}
\end{figure*}

\textbf{Relative Clustered Distance Maps}.
We present an extension of Fig.~\ref{fig:heatmap} for both the inputs, $\boldsymbol{\xi}$ \& $\boldsymbol{x}$, and outputs, $\boldsymbol{\beta}$ \& $\boldsymbol{\eta}$, of the \metanets~in Fig.~\ref{fig:app_heatmap}.

\begin{figure}[!t]
\centering

\begin{subfigure}{\columnwidth}
    \includegraphics[trim=0 0 0 0, clip, width=0.9\columnwidth]{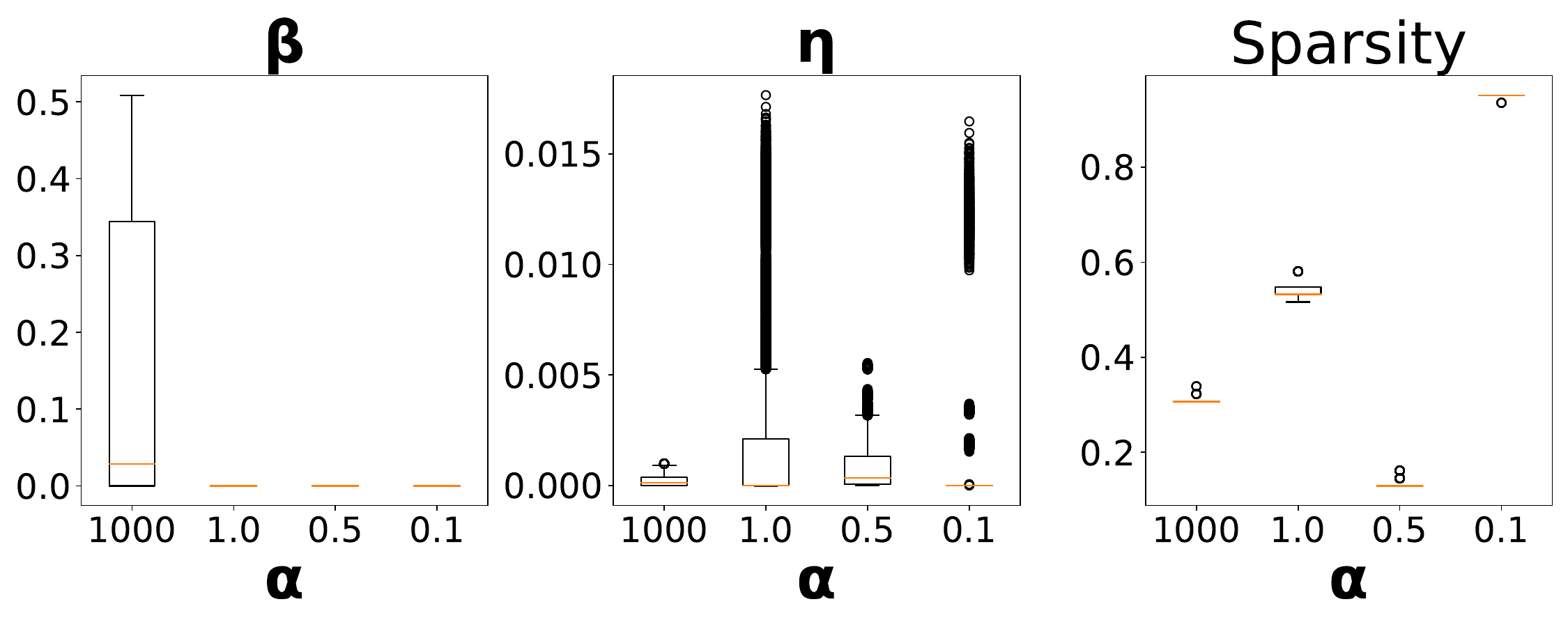}
    
    \vspace{-5pt}
    \caption{CIFAR10 ($e=5$)}
\end{subfigure}
\vspace{-1pt}
\begin{subfigure}{\columnwidth}
    \includegraphics[trim=0 0pt 0 0pt, clip,width=0.9\columnwidth]{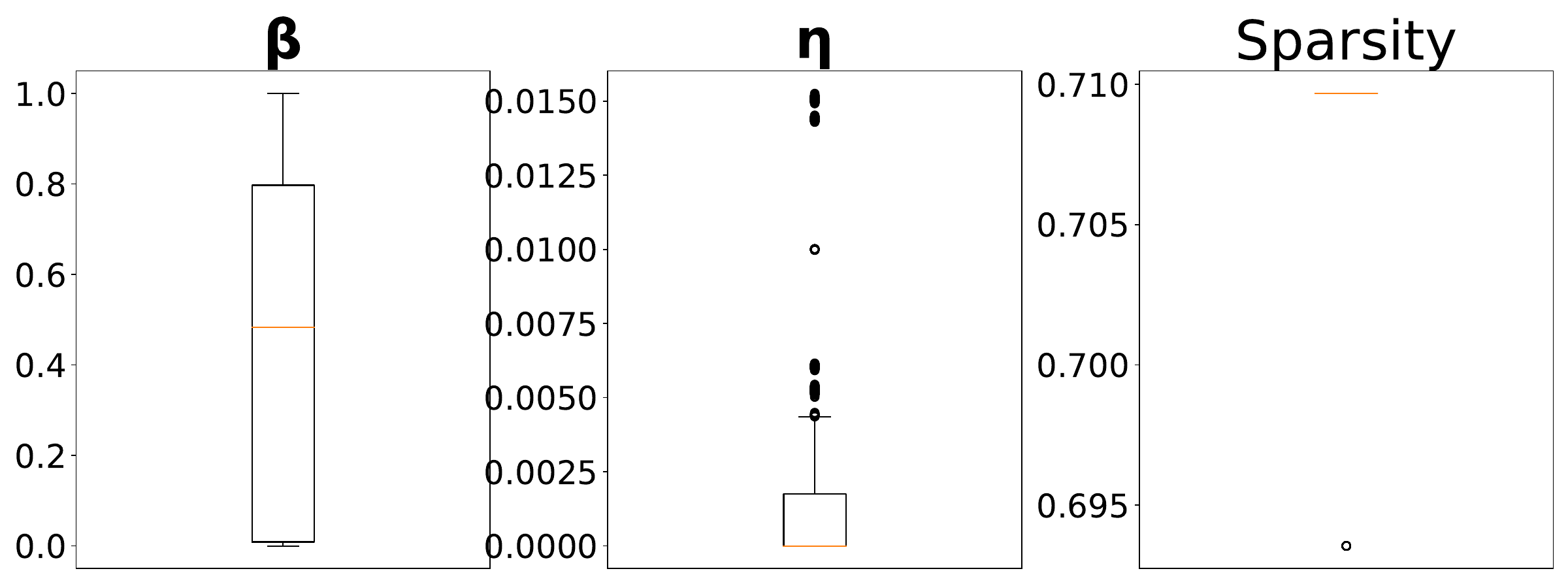}
    
    \vspace{-5pt}
    \caption{Speech Commands}
    \label{fig:hp_commands}
\end{subfigure}
        
\caption{Locality, spread, and skewness of \method's learned hyperparameters, $\boldsymbol{\beta}$ and $\boldsymbol{\eta}$, of different layers across clients and the model's sparsity of all clients for each personalization scenario.}
\label{fig:app_hp}
\end{figure}

\textbf{Learned Personalized Hyperparameters}.
We present the learned hyperparameters for the other client groups not shown in Fig.~\ref{fig:hp} in Fig.~\ref{fig:app_hp}.

\textbf{Comparison with FOMAML}.
We remark that off-the-shelf FOMAML~\cite{maml} (as an initial condition learner) is not a meaningful point of comparison because our problem is to fine-tune/personalize pre-trained models. Therefore, to compare with FOMAML, we focus on learning our \method{} meta-nets with FOMAML style meta-gradients instead of IFT meta-gradients. In order to apply FOMAML to learning rate optimization, this also required extending FOMAML with the same trick as we did for our IFT approach as shown in Eq.~\ref{eq:ift_loss_train-1} \& ~\ref{eq:ift_loss_train-2}. However, FOMAML ignores the learning trajectory except the last step, which may result in performance degradation over longer horizons. We term this baseline FOMAML+. Table.~\ref{tab:domain_fomaml} reports results on the multi-domain datasets as described in Section.~\ref{subsec:officedomainnet}, keeping the same initial condition and meta representation (LRNet and BNNet meta-nets), and varying only the optimization algorithm. Our approach of using IFT to compute the best response Jacobian, outperforms FOMAML+ for $\alpha=1000,1.0,0.5$ and has comparable performance for $\alpha=0.1$.

\input{tables/domain_fomaml}

%% file: tables/domain_fomaml.tex
\begin{table}[!t]
\centering
\caption{Comparison between using FOMAML+ and IFT in \method{} for Office-Caltech-10 and Domainnet ($e=15$).}
\label{tab:domain_fomaml}
\vspace{0.2em}
\begin{scriptsize}
\resizebox{1.0\textwidth}{!}{\begin{minipage}{\textwidth}
\begin{center}
\begin{tabular}{|ccccc|}
\hline

\multicolumn{1}{|c|}{\textbf{$\boldsymbol{\alpha}$}} &
\multicolumn{1}{|c|}{\textbf{Dataset}} &
  \multicolumn{1}{c|}{\textbf{+FT (BN C)}} &
  \multicolumn{1}{c|}{\textbf{+\method{} (FOMAML+)}} &
  \textbf{+\method{} (IFT)} \\ \hline
\multicolumn{1}{|c|}{\textbf{1000}} &
\multicolumn{1}{|c|}{Caltech-10} &
  \multicolumn{1}{c|}{80.97±0.33} &
  \multicolumn{1}{c|}{83.20±1.92} &
  \multicolumn{1}{c|}{\textbf{88.85±0.89}} \\
  \multicolumn{1}{|c|}{} &
  \multicolumn{1}{|c|}{DomainNet} &
  \multicolumn{1}{c|}{52.17±1.55} &
  \multicolumn{1}{c|}{52.70±0.17} &
  \multicolumn{1}{c|}{\textbf{54.38±0.45}} \\ \hline
\multicolumn{1}{|c|}{\textbf{1.0}} &
\multicolumn{1}{|c|}{DomainNet} &
  \multicolumn{1}{c|}{62.27±0.58} &
  \multicolumn{1}{c|}{63.14±0.38} &
  \textbf{63.77±0.44} \\ \hline
\multicolumn{1}{|c|}{\textbf{0.5}} &
\multicolumn{1}{|c|}{DomainNet} &
  \multicolumn{1}{c|}{71.39±0.97} &
  \multicolumn{1}{c|}{71.37±0.79} &
  \textbf{72.64±0.3} \\ \hline
\multicolumn{1}{|c|}{\textbf{0.1}} &
\multicolumn{1}{|c|}{DomainNet} &
  \multicolumn{1}{c|}{86.03±0.47} &
  \multicolumn{1}{c|}{86.22±0.16} &
  \textbf{86.36±0.45} \\ \hline
\end{tabular}
\end{center}
\end{minipage}
}
\end{scriptsize}
\end{table}

%% file: main.bbl
\begin{thebibliography}{65}
\providecommand{\natexlab}[1]{#1}
\providecommand{\url}[1]{\texttt{#1}}
\expandafter\ifx\csname urlstyle\endcsname\relax
  \providecommand{\doi}[1]{doi: #1}\else
  \providecommand{\doi}{doi: \begingroup \urlstyle{rm}\Url}\fi

\bibitem[Agarap(2018)]{agarap2018deep}
Abien~Fred Agarap.
\newblock Deep learning using rectified linear units (relu).
\newblock \emph{arXiv preprint arXiv:1803.08375}, 2018.

\bibitem[Arivazhagan et~al.(2019)Arivazhagan, Aggarwal, Singh, and
  Choudhary]{fedper}
Manoj~Ghuhan Arivazhagan, Vinay Aggarwal, Aaditya~Kumar Singh, and Sunav
  Choudhary.
\newblock Federated learning with personalization layers, 2019.

\bibitem[Baik et~al.(2020)Baik, Choi, Choi, Kim, and Lee]{baik2020meta}
Sungyong Baik, Myungsub Choi, Janghoon Choi, Heewon Kim, and Kyoung~Mu Lee.
\newblock Meta-learning with adaptive hyperparameters.
\newblock \emph{Advances in Neural Information Processing Systems},
  33:\penalty0 20755--20765, 2020.

\bibitem[Bengio et~al.(2013)Bengio, L{\'e}onard, and
  Courville]{bengio2013estimating}
Yoshua Bengio, Nicholas L{\'e}onard, and Aaron Courville.
\newblock Estimating or propagating gradients through stochastic neurons for
  conditional computation.
\newblock \emph{arXiv preprint arXiv:1308.3432}, 2013.

\bibitem[Bernacchia(2021)]{bernacchia2021metalearning}
Alberto Bernacchia.
\newblock Meta-learning with negative learning rates.
\newblock In \emph{International Conference on Learning Representations}, 2021.

\bibitem[Beutel et~al.(2020)Beutel, Topal, Mathur, Qiu, Parcollet,
  de~Gusm{\~a}o, and Lane]{flower}
Daniel~J Beutel, Taner Topal, Akhil Mathur, Xinchi Qiu, Titouan Parcollet,
  Pedro~PB de~Gusm{\~a}o, and Nicholas~D Lane.
\newblock Flower: A friendly federated learning research framework.
\newblock \emph{arXiv preprint arXiv:2007.14390}, 2020.

\bibitem[Briggs et~al.(2020)Briggs, Fan, and Andras]{flhc}
Christopher Briggs, Zhong Fan, and Peter Andras.
\newblock Federated learning with hierarchical clustering of local updates to
  improve training on non-iid data.
\newblock In \emph{2020 International Joint Conference on Neural Networks
  (IJCNN)}. IEEE, 2020.

\bibitem[Chang et~al.(2019)Chang, You, Seo, Kwak, and Han]{chang2019domain}
Woong-Gi Chang, Tackgeun You, Seonguk Seo, Suha Kwak, and Bohyung Han.
\newblock Domain-specific batch normalization for unsupervised domain
  adaptation.
\newblock In \emph{Proceedings of the IEEE/CVF conference on Computer Vision
  and Pattern Recognition (CVPR)}, pages 7354--7362, 2019.

\bibitem[Chen et~al.(2022)Chen, Gao, Kuang, Li, and Ding]{chen2022pfl}
Daoyuan Chen, Dawei Gao, Weirui Kuang, Yaliang Li, and Bolin Ding.
\newblock pfl-bench: A comprehensive benchmark for personalized federated
  learning.
\newblock \emph{Advances in Neural Information Processing Systems}, 2022.

\bibitem[Chen et~al.(2018)Chen, Luo, Dong, Li, and He]{fedmeta}
Fei Chen, Mi~Luo, Zhenhua Dong, Zhenguo Li, and Xiuqiang He.
\newblock Federated meta-learning with fast convergence and efficient
  communication.
\newblock \emph{arXiv preprint arXiv:1802.07876}, 2018.

\bibitem[Collins et~al.(2021)Collins, Hassani, Mokhtari, and
  Shakkottai]{fedrep}
Liam Collins, Hamed Hassani, Aryan Mokhtari, and Sanjay Shakkottai.
\newblock Exploiting shared representations for personalized federated
  learning.
\newblock In \emph{International Conference on Machine Learning}. PMLR, 2021.

\bibitem[Cs{\'a}ji et~al.(2001)]{csaji2001approximation}
Bal{\'a}zs~Csan{\'a}d Cs{\'a}ji et~al.
\newblock Approximation with artificial neural networks.
\newblock \emph{Faculty of Sciences, Etvs Lornd University, Hungary},
  24\penalty0 (48):\penalty0 7, 2001.

\bibitem[Deng et~al.(2009)Deng, Dong, Socher, Li, Li, and Fei-Fei]{imagenet}
Jia Deng, Wei Dong, Richard Socher, Li-Jia Li, Kai Li, and Li~Fei-Fei.
\newblock Imagenet: A large-scale hierarchical image database.
\newblock In \emph{2009 IEEE Conference on Computer Vision and Pattern
  Recognition}, pages 248--255. IEEE, 2009.

\bibitem[Deng et~al.(2020)Deng, Kamani, and Mahdavi]{apfl}
Yuyang Deng, Mohammad~Mahdi Kamani, and Mehrdad Mahdavi.
\newblock Adaptive personalized federated learning.
\newblock \emph{arXiv preprint arXiv:2003.13461}, 2020.

\bibitem[Fallah et~al.(2020)Fallah, Mokhtari, and Ozdaglar]{perfedavg}
Alireza Fallah, Aryan Mokhtari, and Asuman Ozdaglar.
\newblock Personalized federated learning with theoretical guarantees: A
  model-agnostic meta-learning approach.
\newblock \emph{Advances in Neural Information Processing Systems}, 2020.

\bibitem[Finn et~al.(2017)Finn, Abbeel, and Levine]{maml}
Chelsea Finn, Pieter Abbeel, and Sergey Levine.
\newblock Model-agnostic meta-learning for fast adaptation of deep networks.
\newblock In \emph{International Conference on Machine Learning}. PMLR, 2017.

\bibitem[Ghosh et~al.(2020)Ghosh, Chung, Yin, and Ramchandran]{ifca}
Avishek Ghosh, Jichan Chung, Dong Yin, and Kannan Ramchandran.
\newblock An efficient framework for clustered federated learning.
\newblock \emph{Advances in Neural Information Processing Systems}, 2020.

\bibitem[Glorot and Bengio(2010)]{glorot2010understanding}
Xavier Glorot and Yoshua Bengio.
\newblock Understanding the difficulty of training deep feedforward neural
  networks.
\newblock In \emph{Proceedings of the thirteenth international conference on
  artificial intelligence and statistics}, pages 249--256. JMLR Workshop and
  Conference Proceedings, 2010.

\bibitem[Gong et~al.(2012)Gong, Shi, Sha, and Grauman]{office}
Boqing Gong, Yuan Shi, Fei Sha, and Kristen Grauman.
\newblock Geodesic flow kernel for unsupervised domain adaptation.
\newblock In \emph{2012 IEEE Conference on Computer Vision and Pattern
  Recognition}, pages 2066--2073. IEEE, 2012.

\bibitem[He et~al.(2016)He, Zhang, Ren, and Sun]{resnet}
Kaiming He, Xiangyu Zhang, Shaoqing Ren, and Jian Sun.
\newblock Deep residual learning for image recognition.
\newblock In \emph{Proceedings of the IEEE conference on computer vision and
  pattern recognition}, pages 770--778, 2016.

\bibitem[Hendrycks and Dietterich(2019)]{cifar10c}
Dan Hendrycks and Thomas Dietterich.
\newblock Benchmarking neural network robustness to common corruptions and
  perturbations.
\newblock \emph{Proceedings of the International Conference on Learning
  Representations}, 2019.

\bibitem[Holly et~al.(2022)Holly, Hiessl, Lakani, Schall, Heitzinger, and
  Kemnitz]{holly2022evaluation}
Stephanie Holly, Thomas Hiessl, Safoura~Rezapour Lakani, Daniel Schall, Clemens
  Heitzinger, and Jana Kemnitz.
\newblock Evaluation of hyperparameter-optimization approaches in an industrial
  federated learning system.
\newblock In \emph{Data Science--Analytics and Applications}. Springer, 2022.

\bibitem[Horvath et~al.(2021)Horvath, Laskaridis, Almeida, Leontiadis,
  Venieris, and Lane]{fjord}
Samuel Horvath, Stefanos Laskaridis, Mario Almeida, Ilias Leontiadis,
  Stylianos~I Venieris, and Nicholas~D Lane.
\newblock Fjord: Fair and accurate federated learning under heterogeneous
  targets with ordered dropout.
\newblock \emph{arXiv preprint arXiv:2102.13451}, 2021.

\bibitem[Hsu et~al.(2019)Hsu, Qi, and Brown]{hsu2019measuring}
Tzu-Ming~Harry Hsu, Hang Qi, and Matthew Brown.
\newblock Measuring the effects of non-identical data distribution for
  federated visual classification.
\newblock \emph{arXiv preprint arXiv:1909.06335}, 2019.

\bibitem[Hubert and Arabie(1985)]{ari}
Lawrence Hubert and Phipps Arabie.
\newblock Comparing partitions.
\newblock \emph{Journal of Classification}, 2\penalty0 (1):\penalty0 193--218,
  1985.

\bibitem[Ioffe and Szegedy(2015)]{ioffe2015batch}
Sergey Ioffe and Christian Szegedy.
\newblock Batch normalization: Accelerating deep network training by reducing
  internal covariate shift.
\newblock In \emph{International conference on machine learning}, pages
  448--456. PMLR, 2015.

\bibitem[Jiang et~al.(2019)Jiang, Kone{\v{c}}n{\`y}, Rush, and
  Kannan]{jiang2019improving}
Yihan Jiang, Jakub Kone{\v{c}}n{\`y}, Keith Rush, and Sreeram Kannan.
\newblock Improving federated learning personalization via model agnostic meta
  learning.
\newblock \emph{arXiv preprint arXiv:1909.12488}, 2019.

\bibitem[Kairouz et~al.(2021)Kairouz, McMahan, Avent, Bellet, Bennis, Bhagoji,
  Bonawitz, Charles, Cormode, Cummings, et~al.]{kairouz2021advances}
Peter Kairouz, H~Brendan McMahan, Brendan Avent, Aur{\'e}lien Bellet, Mehdi
  Bennis, Arjun~Nitin Bhagoji, Kallista Bonawitz, Zachary Charles, Graham
  Cormode, Rachel Cummings, et~al.
\newblock Advances and open problems in federated learning.
\newblock \emph{Foundations and Trends in Machine Learning}, 14\penalty0
  (1--2):\penalty0 1--210, 2021.

\bibitem[Karimireddy et~al.(2020)Karimireddy, Kale, Mohri, Reddi, Stich, and
  Suresh]{scaffold}
Sai~Praneeth Karimireddy, Satyen Kale, Mehryar Mohri, Sashank Reddi, Sebastian
  Stich, and Ananda~Theertha Suresh.
\newblock {SCAFFOLD}: Stochastic controlled averaging for federated learning.
\newblock In \emph{Proceedings of the 37th International Conference on Machine
  Learning}. PMLR, 2020.

\bibitem[Khodak et~al.(2021)Khodak, Tu, Li, Li, Balcan, Smith, and
  Talwalkar]{fedex}
Mikhail Khodak, Renbo Tu, Tian Li, Liam Li, Maria-Florina~F Balcan, Virginia
  Smith, and Ameet Talwalkar.
\newblock Federated hyperparameter tuning: Challenges, baselines, and
  connections to weight-sharing.
\newblock \emph{Advances in Neural Information Processing Systems}, 34, 2021.

\bibitem[Krizhevsky et~al.(2009)Krizhevsky, Hinton,
  et~al.]{krizhevsky2009learning}
Alex Krizhevsky, Geoffrey Hinton, et~al.
\newblock Learning multiple layers of features from tiny images.
\newblock 2009.

\bibitem[Li et~al.(2021{\natexlab{a}})Li, He, and Song]{moon}
Qinbin Li, Bingsheng He, and Dawn Song.
\newblock Model-contrastive federated learning.
\newblock In \emph{Proceedings of the IEEE/CVF Conference on Computer Vision
  and Pattern Recognition}, 2021{\natexlab{a}}.

\bibitem[Li et~al.(2020)Li, Sahu, Zaheer, Sanjabi, Talwalkar, and
  Smith]{fedprox}
Tian Li, Anit~Kumar Sahu, Manzil Zaheer, Maziar Sanjabi, Ameet Talwalkar, and
  Virginia Smith.
\newblock Federated optimization in heterogeneous networks.
\newblock \emph{Proceedings of Machine Learning and Systems}, 2:\penalty0
  429--450, 2020.

\bibitem[Li et~al.(2021{\natexlab{b}})Li, Hu, Beirami, and Smith]{ditto}
Tian Li, Shengyuan Hu, Ahmad Beirami, and Virginia Smith.
\newblock Ditto: Fair and robust federated learning through personalization.
\newblock In \emph{International Conference on Machine Learning}. PMLR,
  2021{\natexlab{b}}.

\bibitem[Li et~al.(2021{\natexlab{c}})Li, Jiang, Zhang, Kamp, and Dou]{fedbn}
Xiaoxiao Li, Meirui Jiang, Xiaofei Zhang, Michael Kamp, and Qi~Dou.
\newblock Fedbn: Federated learning on non-iid features via local batch
  normalization.
\newblock \emph{arXiv preprint arXiv:2102.07623}, 2021{\natexlab{c}}.

\bibitem[Li et~al.(2016)Li, Wang, Shi, Liu, and Hou]{li2016revisiting}
Yanghao Li, Naiyan Wang, Jianping Shi, Jiaying Liu, and Xiaodi Hou.
\newblock Revisiting batch normalization for practical domain adaptation.
\newblock \emph{arXiv preprint arXiv:1603.04779}, 2016.

\bibitem[Liang et~al.(2020)Liang, Liu, Ziyin, Allen, Auerbach, Brent,
  Salakhutdinov, and Morency]{lgfedavg}
Paul~Pu Liang, Terrance Liu, Liu Ziyin, Nicholas~B Allen, Randy~P Auerbach,
  David Brent, Ruslan Salakhutdinov, and Louis-Philippe Morency.
\newblock Think locally, act globally: Federated learning with local and global
  representations.
\newblock \emph{arXiv preprint arXiv:2001.01523}, 2020.

\bibitem[Lorraine et~al.(2020)Lorraine, Vicol, and
  Duvenaud]{lorraine2020optimizing}
Jonathan Lorraine, Paul Vicol, and David Duvenaud.
\newblock Optimizing millions of hyperparameters by implicit differentiation.
\newblock In \emph{International Conference on Artificial Intelligence and
  Statistics}, pages 1540--1552. PMLR, 2020.

\bibitem[Luo et~al.(2021)Luo, Chen, Hu, Zhang, Liang, and Feng]{ccvr}
Mi~Luo, Fei Chen, Dapeng Hu, Yifan Zhang, Jian Liang, and Jiashi Feng.
\newblock No fear of heterogeneity: Classifier calibration for federated
  learning with non-iid data.
\newblock \emph{Advances in Neural Information Processing Systems}, 2021.

\bibitem[Ma et~al.(2022)Ma, Zhang, Guo, and Xu]{pFedLA}
Xiaosong Ma, Jie Zhang, Song Guo, and Wenchao Xu.
\newblock Layer-wised model aggregation for personalized federated learning.
\newblock In \emph{Proceedings of the IEEE/CVF Conference on Computer Vision
  and Pattern Recognition}, 2022.

\bibitem[maintainers and contributors(2016)]{torchvision}
TorchVision maintainers and contributors.
\newblock Torchvision: Pytorch's computer vision library.
\newblock \url{https://github.com/pytorch/vision}, 2016.

\bibitem[Mansour et~al.(2020)Mansour, Mohri, Ro, and Suresh]{hypcluster}
Yishay Mansour, Mehryar Mohri, Jae Ro, and Ananda~Theertha Suresh.
\newblock Three approaches for personalization with applications to federated
  learning.
\newblock \emph{arXiv preprint arXiv:2002.10619}, 2020.

\bibitem[Marfoq et~al.(2021)Marfoq, Neglia, Bellet, Kameni, and Vidal]{fedem}
Othmane Marfoq, Giovanni Neglia, Aur{\'e}lien Bellet, Laetitia Kameni, and
  Richard Vidal.
\newblock Federated multi-task learning under a mixture of distributions.
\newblock \emph{Advances in Neural Information Processing Systems}, 2021.

\bibitem[Matsuda et~al.(2022{\natexlab{a}})Matsuda, Sasaki, Xiao, and
  Onizuka]{fedme}
Koji Matsuda, Yuya Sasaki, Chuan Xiao, and Makoto Onizuka.
\newblock Fedme: Federated learning via model exchange.
\newblock In \emph{Proceedings of the 2022 SIAM International Conference on
  Data Mining (SDM)}. SIAM, 2022{\natexlab{a}}.

\bibitem[Matsuda et~al.(2022{\natexlab{b}})Matsuda, Sasaki, Xiao, and
  Onizuka]{matsuda2022empirical}
Koji Matsuda, Yuya Sasaki, Chuan Xiao, and Makoto Onizuka.
\newblock An empirical study of personalized federated learning.
\newblock \emph{arXiv preprint arXiv:2206.13190}, 2022{\natexlab{b}}.

\bibitem[McMahan et~al.(2017)McMahan, Moore, Ramage, Hampson, and
  y~Arcas]{fedavg}
Brendan McMahan, Eider Moore, Daniel Ramage, Seth Hampson, and Blaise~Aguera
  y~Arcas.
\newblock Communication-efficient learning of deep networks from decentralized
  data.
\newblock In \emph{Artificial intelligence and statistics}. PMLR, 2017.

\bibitem[Navon et~al.(2021)Navon, Achituve, Maron, Chechik, and
  Fetaya]{navon2021auxiliary}
Aviv Navon, Idan Achituve, Haggai Maron, Gal Chechik, and Ethan Fetaya.
\newblock Auxiliary learning by implicit differentiation.
\newblock In \emph{International Conference on Learning Representations}, 2021.

\bibitem[Oh et~al.(2021)Oh, Kim, and Yun]{fedbabu}
Jaehoon Oh, Sangmook Kim, and Se-Young Yun.
\newblock Fedbabu: Towards enhanced representation for federated image
  classification.
\newblock \emph{arXiv preprint arXiv:2106.06042}, 2021.

\bibitem[Oreshkin et~al.(2018)Oreshkin, Rodr{\'\i}guez~L{\'o}pez, and
  Lacoste]{oreshkin2018tadam}
Boris Oreshkin, Pau Rodr{\'\i}guez~L{\'o}pez, and Alexandre Lacoste.
\newblock Tadam: Task dependent adaptive metric for improved few-shot learning.
\newblock \emph{Advances in neural information processing systems}, 31, 2018.

\bibitem[Pedregosa et~al.(2011)Pedregosa, Varoquaux, Gramfort, Michel, Thirion,
  Grisel, Blondel, Prettenhofer, Weiss, Dubourg, Vanderplas, Passos,
  Cournapeau, Brucher, Perrot, and Duchesnay]{scikit-learn}
F.~Pedregosa, G.~Varoquaux, A.~Gramfort, V.~Michel, B.~Thirion, O.~Grisel,
  M.~Blondel, P.~Prettenhofer, R.~Weiss, V.~Dubourg, J.~Vanderplas, A.~Passos,
  D.~Cournapeau, M.~Brucher, M.~Perrot, and E.~Duchesnay.
\newblock Scikit-learn: Machine learning in {P}ython.
\newblock \emph{Journal of Machine Learning Research}, 12:\penalty0 2825--2830,
  2011.

\bibitem[Peng et~al.(2019)Peng, Bai, Xia, Huang, Saenko, and Wang]{domainnet}
Xingchao Peng, Qinxun Bai, Xide Xia, Zijun Huang, Kate Saenko, and Bo~Wang.
\newblock Moment matching for multi-source domain adaptation.
\newblock In \emph{Proceedings of the IEEE/CVF International Conference on
  Computer Vision}, pages 1406--1415, 2019.

\bibitem[Qiu et~al.(2021)Qiu, Fernandez-Marques, Gusmao, Gao, Parcollet, and
  Lane]{zerofl}
Xinchi Qiu, Javier Fernandez-Marques, Pedro~PB Gusmao, Yan Gao, Titouan
  Parcollet, and Nicholas~Donald Lane.
\newblock Zerofl: Efficient on-device training for federated learning with
  local sparsity.
\newblock In \emph{International Conference on Learning Representations}, 2021.

\bibitem[Sattler et~al.(2020)Sattler, M{\"u}ller, and Samek]{cfl}
Felix Sattler, Klaus-Robert M{\"u}ller, and Wojciech Samek.
\newblock Clustered federated learning: Model-agnostic distributed multitask
  optimization under privacy constraints.
\newblock \emph{IEEE Transactions on Neural Networks and Learning Systems},
  2020.

\bibitem[Setayesh et~al.()Setayesh, Li, and Wong]{perfedmask}
Mehdi Setayesh, Xiaoxiao Li, and Vincent~WS Wong.
\newblock Perfedmask: Personalized federated learning with optimized masking
  vectors.
\newblock In \emph{The Eleventh International Conference on Learning
  Representations}.

\bibitem[Shamsian et~al.(2021)Shamsian, Navon, Fetaya, and Chechik]{pfedhn}
Aviv Shamsian, Aviv Navon, Ethan Fetaya, and Gal Chechik.
\newblock Personalized federated learning using hypernetworks.
\newblock In \emph{International Conference on Machine Learning}, 2021.

\bibitem[Shen et~al.(2020)Shen, Zhang, Jia, Zhang, Huang, Zhou, Kuang, Wu, and
  Wu]{fml}
Tao Shen, Jie Zhang, Xinkang Jia, Fengda Zhang, Gang Huang, Pan Zhou, Kun
  Kuang, Fei Wu, and Chao Wu.
\newblock Federated mutual learning.
\newblock \emph{arXiv preprint arXiv:2006.16765}, 2020.

\bibitem[Shu et~al.(2019)Shu, Xie, Yi, Zhao, Zhou, Xu, and
  Meng]{meta_weight_net}
Jun Shu, Qi~Xie, Lixuan Yi, Qian Zhao, Sanping Zhou, Zongben Xu, and Deyu Meng.
\newblock Meta-weight-net: Learning an explicit mapping for sample weighting.
\newblock \emph{Advances in neural information processing systems}, 32, 2019.

\bibitem[Stella and Shi(2003)]{stella2003multiclass}
X~Yu Stella and Jianbo Shi.
\newblock Multiclass spectral clustering.
\newblock In \emph{IEEE International Conference on Computer Vision}, volume~2,
  pages 313--313. IEEE Computer Society, 2003.

\bibitem[T~Dinh et~al.(2020)T~Dinh, Tran, and Nguyen]{pfedme}
Canh T~Dinh, Nguyen Tran, and Josh Nguyen.
\newblock Personalized federated learning with moreau envelopes.
\newblock \emph{Advances in Neural Information Processing Systems},
  33:\penalty0 21394--21405, 2020.

\bibitem[Warden(2018)]{warden2018speech}
Pete Warden.
\newblock Speech commands: A dataset for limited-vocabulary speech recognition.
\newblock \emph{arXiv preprint arXiv:1804.03209}, 2018.

\bibitem[Yurochkin et~al.(2019)Yurochkin, Agarwal, Ghosh, Greenewald, Hoang,
  and Khazaeni]{yurochkin2019bayesian}
Mikhail Yurochkin, Mayank Agarwal, Soumya Ghosh, Kristjan Greenewald, Nghia
  Hoang, and Yasaman Khazaeni.
\newblock Bayesian nonparametric federated learning of neural networks.
\newblock In \emph{International Conference on Machine Learning}, pages
  7252--7261. PMLR, 2019.

\bibitem[Zhang et~al.(2021)Zhang, Sapra, Fidler, Yeung, and Alvarez]{fedfomo}
Michael Zhang, Karan Sapra, Sanja Fidler, Serena Yeung, and Jose~M Alvarez.
\newblock Personalized federated learning with first order model optimization.
\newblock In \emph{International Conference on Learning Representations}, 2021.

\bibitem[Zhao et~al.(2018)Zhao, Li, Lai, Suda, Civin, and Chandra]{emd}
Yue Zhao, Meng Li, Liangzhen Lai, Naveen Suda, Damon Civin, and Vikas Chandra.
\newblock Federated learning with non-iid data.
\newblock \emph{arXiv preprint arXiv:1806.00582}, 2018.

\bibitem[Zhong et~al.(2023)Zhong, He, Ren, Li, and Li]{feddar}
Aoxiao Zhong, Hao He, Zhaolin Ren, Na~Li, and Quanzheng Li.
\newblock Feddar: Federated domain-aware representation learning.
\newblock In \emph{International Conference on Learning Representations}, 2023.

\bibitem[Zhou et~al.(2023)Zhou, Ram, Salonidis, Baracaldo, Samulowitz, and
  Ludwig]{flora}
Yi~Zhou, Parikshit Ram, Theodoros Salonidis, Nathalie Baracaldo, Horst
  Samulowitz, and Heiko Ludwig.
\newblock Single-shot general hyper-parameter optimization for federated
  learning.
\newblock In \emph{The Eleventh International Conference on Learning
  Representations}, 2023.

\end{thebibliography}
